\documentclass[conference]{IEEEtran}

\usepackage{color}
\usepackage{amsmath}
\usepackage{amssymb}
\usepackage{graphicx}
\usepackage{array}
\usepackage{multirow}
\usepackage{graphicx}
\usepackage{times}
\usepackage{epsfig}
\usepackage{mathrsfs}
\usepackage{multirow}
\usepackage{tabularx}
\usepackage{makecell}
\usepackage{array}
\usepackage{algorithm}
\usepackage{algorithmic}
\usepackage[T1]{fontenc}
\usepackage[latin9]{inputenc}
\usepackage{booktabs}
\usepackage{amstext}

\begin{document}
\title{Learning from Synthetic Data \\ Using a Stacked Multichannel Autoencoder}

%\author{
%\IEEEauthorblockN{Xi zhang}
%\IEEEauthorblockA{Illinois Institute of Technology\\
%Chiacgo, IL 60616\\
%Email: xzhang22@hawk.iit.edu}
%\and
%\IEEEauthorblockN{Yanwei Fu}
%\IEEEauthorblockA{Disney Research\\
%Pittsburgh, PA 15213\\
%Email: yanwei.fu@disneyresearch.com }
%\and
%\IEEEauthorblockN{Shanshan Jiang}
%\IEEEauthorblockA{Illinois Institute of Technology\\
%Chiacgo, IL 60616\\
%Email: sjiang20@hawk.iit.edu}
%\and
%\IEEEauthorblockN{Gady Agam}
%\IEEEauthorblockA{Illinois Institute of Technology\\
%Chiacgo, IL 60616\\
%Email: agam@iit.edu}
%}


\author{\IEEEauthorblockN{
Xi Zhang\IEEEauthorrefmark{1},
Yanwei Fu\IEEEauthorrefmark{3},
Shanshan Jiang\IEEEauthorrefmark{2},
Leonid Sigal\IEEEauthorrefmark{3} and
Gady Agam\IEEEauthorrefmark{1}}
\IEEEauthorblockA{
\IEEEauthorrefmark{1}Illinois Institute of Technology,
Chicago, Illinois 60616\\ 
Email: \{xzhang22, sjiang20\}@hawk.iit.edu}
\IEEEauthorblockA{
\IEEEauthorrefmark{2}Illinois Institute of Technology,
Chicago, Illinois 60616\\ 
Email: agam@iit.edu}
\IEEEauthorblockA{
\IEEEauthorrefmark{3}Disney Research, 
Pittsburgh, PA 15213\\
Email: \{yanwei.fu, lsigal\}@disneyresearch.com}
}


\maketitle

\begin{abstract}
%\boldmath
Learning from synthetic data has many important and practical applications, An example of application is photo-sketch recognition. 
Using synthetic data is challenging due to the differences in feature distributions between synthetic and real data, a phenomenon we term {\em synthetic gap}. In this paper, we investigate and formalize 
a general framework -- Stacked Multichannel Autoencoder (SMCAE)  that enables bridging the synthetic 
gap and learning from synthetic data more efficiently. In particular, we show that our SMCAE can not only transform 
and use synthetic data on the challenging face-sketch recognition task, but that it can also help simulate real images, which can be used for training classifiers for recognition. Preliminary experiments validate the effectiveness of the framework.

\end{abstract}


% \fi
\IEEEpeerreviewmaketitle

\section{Introduction}

\begin{figure*}[ht]
\centerline{ %
\begin{tabular}{cccccccc}
\resizebox{0.09\textwidth}{!}{\rotatebox{0}{ \includegraphics{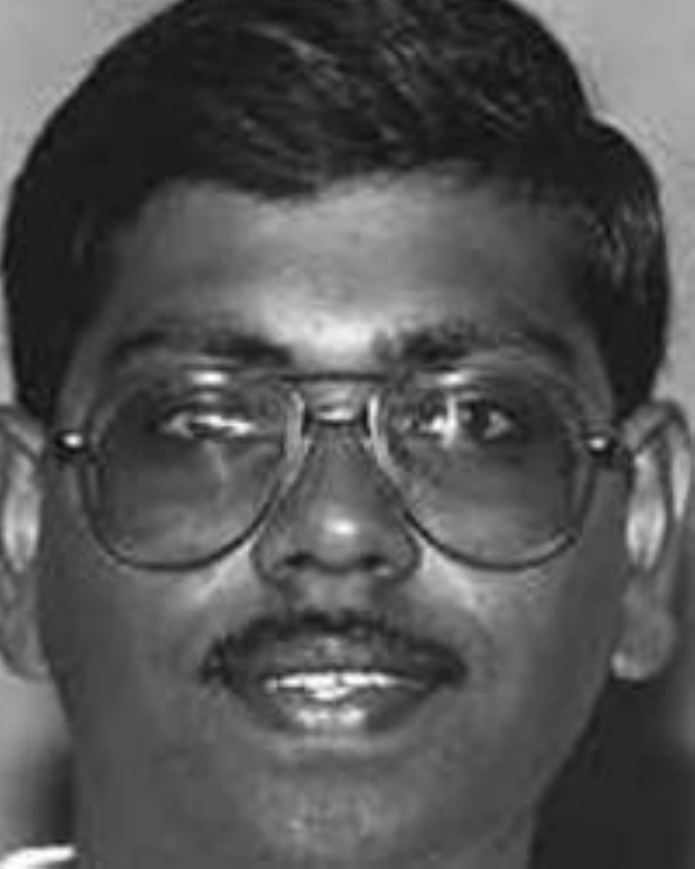}}}  
& 
\resizebox{0.09\textwidth}{!}{\rotatebox{0}{ \includegraphics{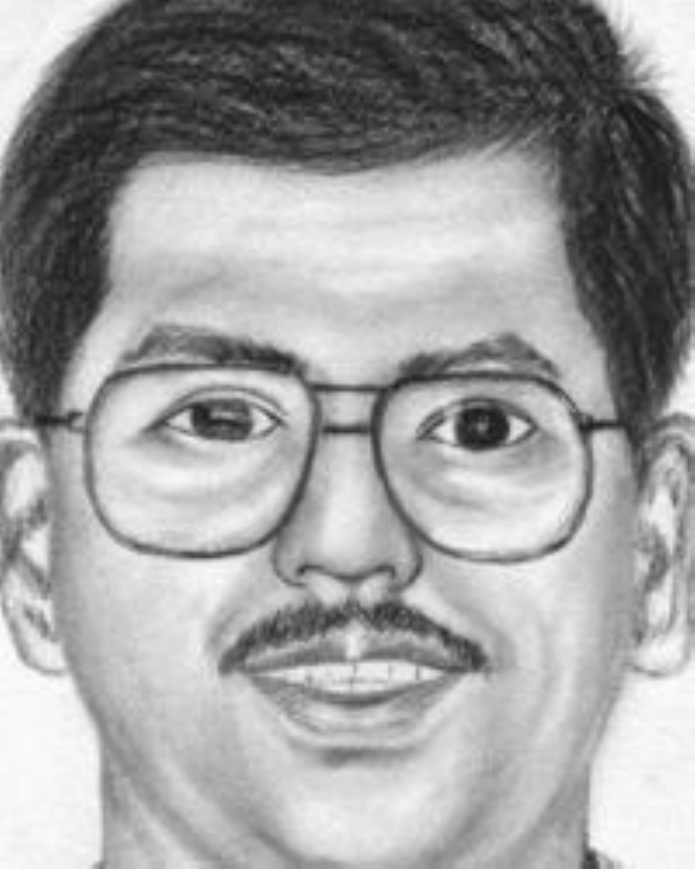}}}  
& 
\resizebox{0.09\textwidth}{!}{\rotatebox{0}{ \includegraphics{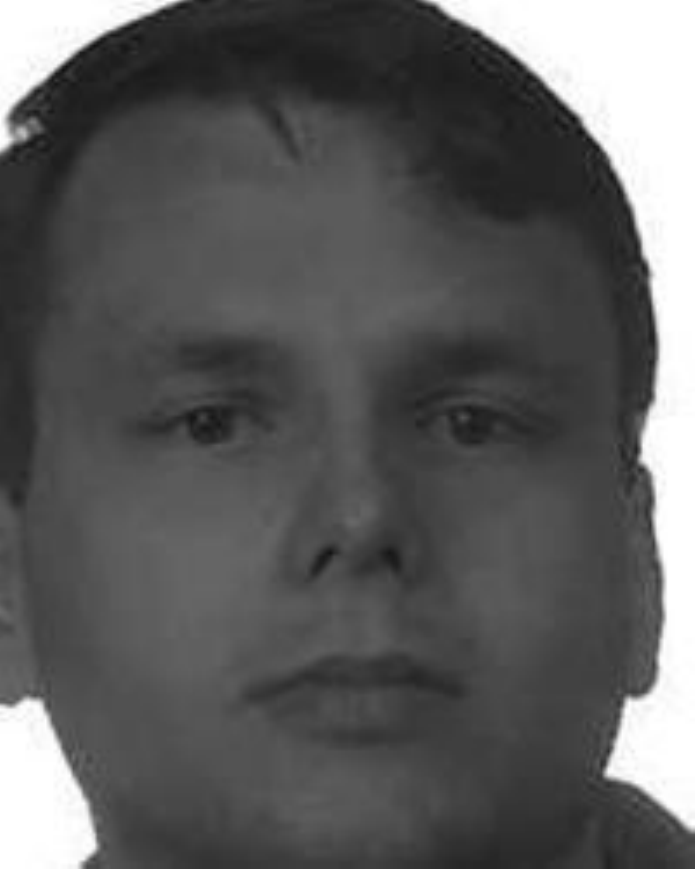}}}  
& 
\resizebox{0.09\textwidth}{!}{\rotatebox{0}{ \includegraphics{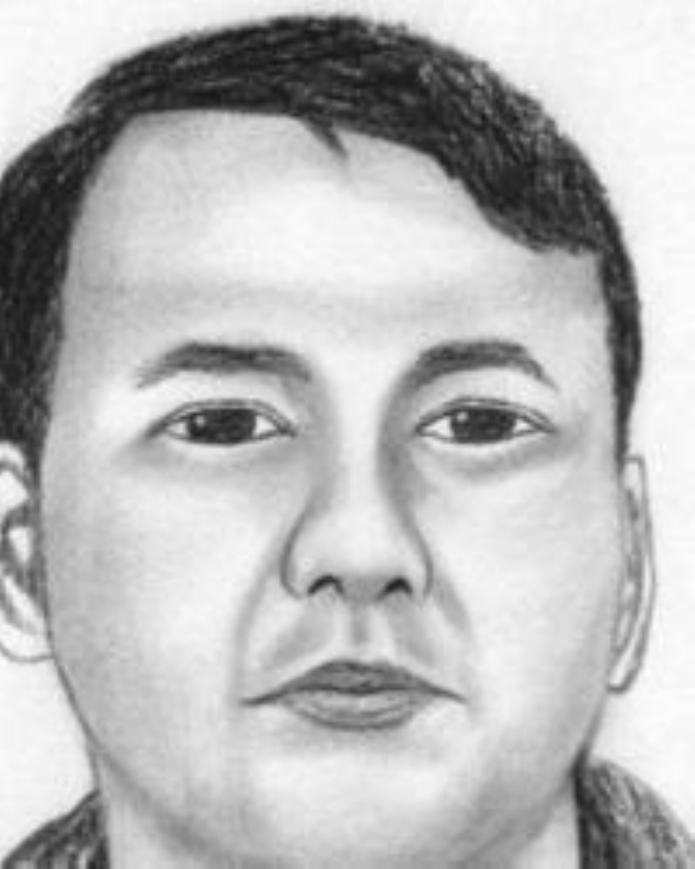}}}  
& 
\resizebox{0.09\textwidth}{!}{\rotatebox{0}{ \includegraphics{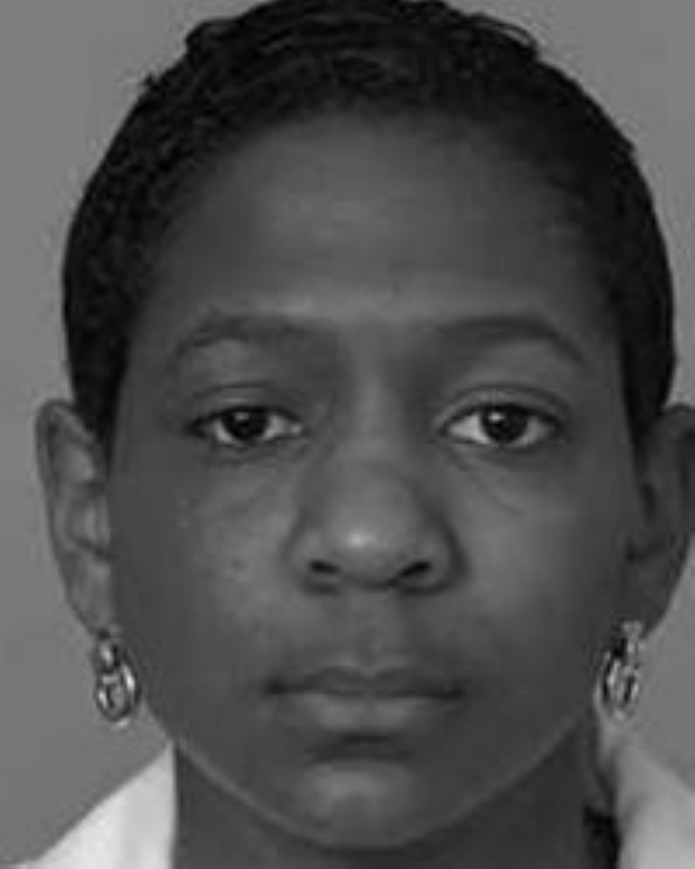}}}  
& 
\resizebox{0.09\textwidth}{!}{\rotatebox{0}{ \includegraphics{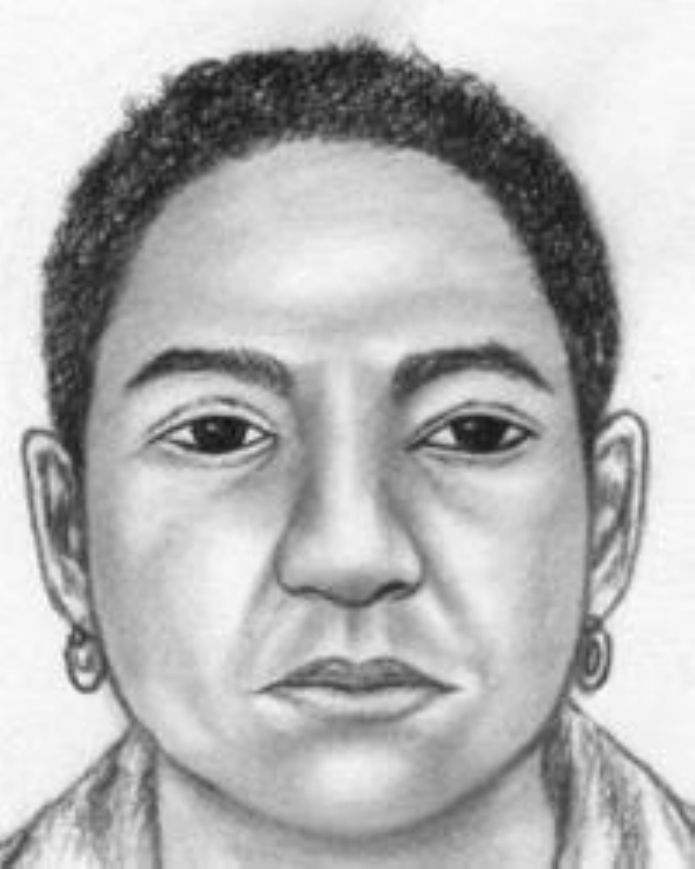}}}  
& 
\resizebox{0.09\textwidth}{!}{\rotatebox{0}{ \includegraphics{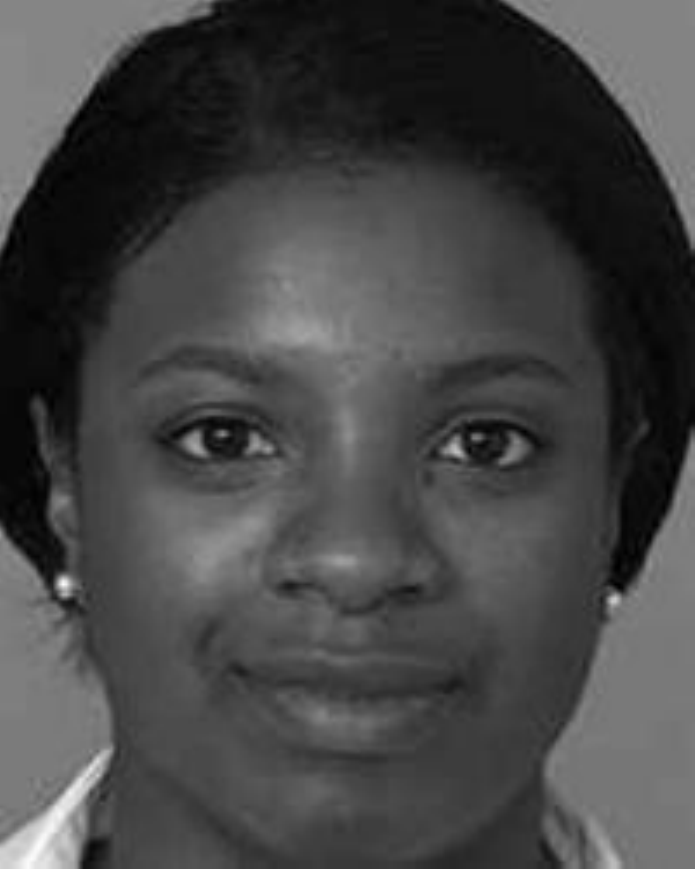}}}  
& 
\resizebox{0.09\textwidth}{!}{\rotatebox{0}{ \includegraphics{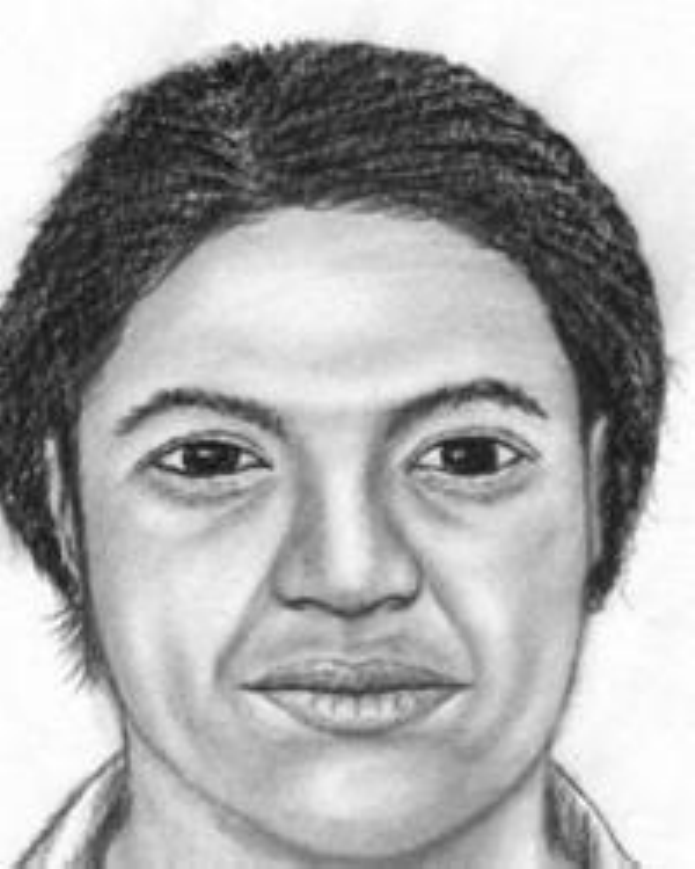}}} \tabularnewline
\end{tabular}} \protect\protect\caption{Examples of face photos and sketches. Data comes from the CUFSF dataset
\cite{wang2009face,zhang2011coupled}.}

\label{fig: feretdemo} 
\end{figure*}

Modern supervised learning algorithms need plenty of data to help train classifiers.  More data with higher quality is always desired in real-world applications; but sometimes, it is beneficial to turn to synthetic data. For example, to help identify criminals, many criminal investigations can only rely on a synthetic face sketch rather than a facial photograph of a suspect which may not be available. Such synthetic
face data is normally drawn by an expert based on descriptions of eyewitnesses and/or victim(s). Several photo-sketch examples are shown in Fig.~\ref{fig: feretdemo}. 
In this application, recognition based on synthetic data is very crucial.

\begin{figure}[ht]
\centerline{ %
\begin{tabular}{cc}
\resizebox{0.24\textwidth}{!}{\rotatebox{0}{ \includegraphics{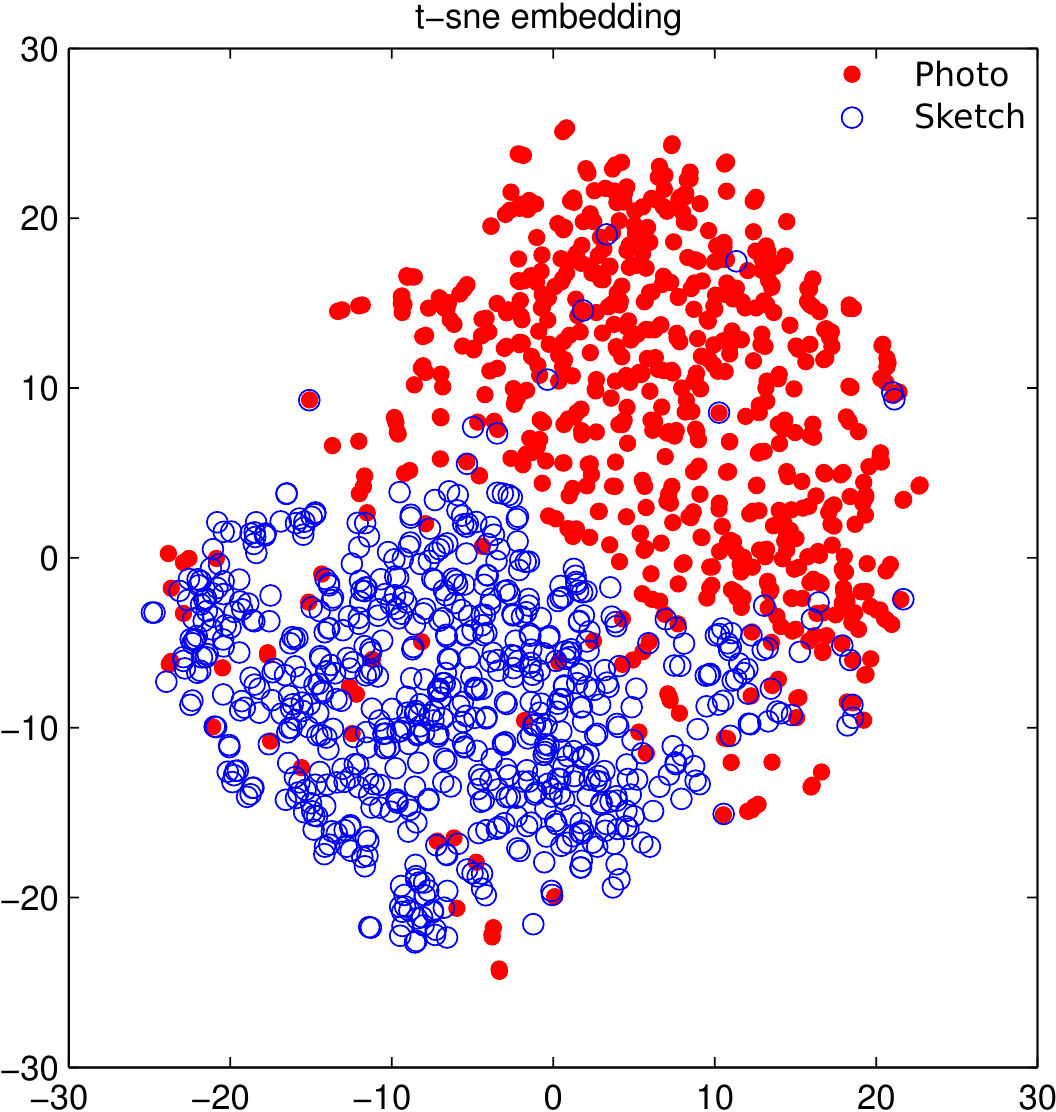}}}  
& \resizebox{0.24\textwidth}{!}{\rotatebox{0}{ \includegraphics{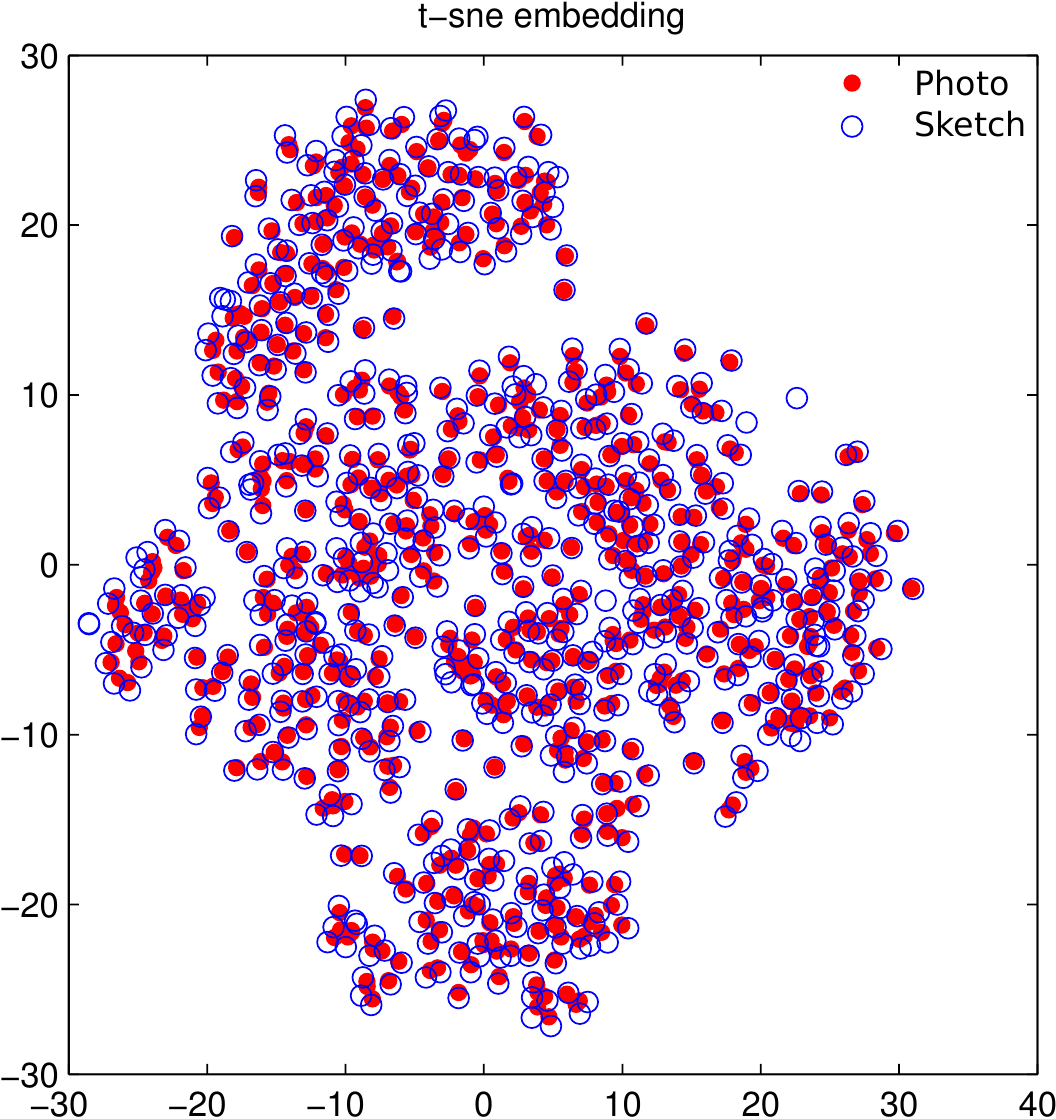}}} \tabularnewline
Original  & Transformed \tabularnewline
\end{tabular}} \protect\caption{\label{fig:t-SNE-visualization-of}t-SNE visualization \cite{van2008visualizing} of the distribution of Histogram of Oriented Gradients (HOG) features in the data in CUFSF dataset \cite{wang2009face,zhang2011coupled}. Left:
synthetic gap is observed between photo and sketch features; Right: the synthetic
gap is bridged by our SMCAE.}
\end{figure}

Directly using synthetic data in a learning algorithm is unfortunately very challenging since synthetic data is different from real data at least to some extent, e.g. exaggerated facial shapes in sketch images in Fig.~\ref{fig: feretdemo} as compared with real images. As a result, the feature distributions of synthetic data may be shifted away from those of real data as illustrated in Fig.~\ref{fig:t-SNE-visualization-of}.
We term such shift in distributions as \emph{synthetic gap}. Synthetic gap is largely caused by the generating process of synthetic data: whereas the synthetic data are generated by replicating principal patterns such as eyes, mouth, nose and hairstyle, rather than replicating every detail of real data. 
The synthetic gap is a major obstacle in using synthetic data in recognition problems, since synthetic data may fail to simulate potentially useful patterns of real data which are important to a successful recognition. 
To solve this problem, we associate synthetic data with real data, and jointly learn from them in a Stacked Multichannel Autoencoder (SMCAE) which can help bridge the synthetic gap by transforming characteristics of synthetic data to better simulate real data.

This paper addresses the problem of learning a mapping from synthetic data to real data. Specifically, we propose a novel framework -- SMCAE. The training process of SMCAE facilitates the bridging of the synthetic gap between the real and the synthetic data by learning how to transform: (1) synthetic to real data and (2) real to real data. In (2) the model learns most essential `characters' and `patterns' of real data, while in (1) it learns how to augment the synthetic data to best reproduce the distribution of real data. Because the two tasks are learned simultaneously, with shared parameters, the essential `characteristics' learned in (2) help to regularize results in (1) and vice versa as we will illustrate in the Handwritten Digit experiments.
% Critically, such mapping, in one task, tries to learn the most essential `characters' and `patterns' of real to real data  which in turn enforce SMCAE in another to learn a transfer from synthetic to real data. Configurations of SMCAE in this way enable the generation of more synthetic data which can best simulate real data by the learned mapping, and we illustrate it in Handwritten Digit experiments.

We highlight two main contributions of this paper: (1) To the best of our knowledge, this is the first attempt to address the problem of {\em synthetic gap}, by demonstrating that the synthetic data could be used to improve the performance on a recognition task. (2) We propose a Stacked Multichannel Autoencoder (SMCAE) model to bridge the {\em synthetic gap} and jointly learn from both real and synthetic data. 

\section{Related Work}

\noindent \textbf{Transfer Learning} aims to extract the knowledge
from one, or more, source tasks and apply it to a target task. 
Transfer learning can be used in many different applications, such as  web page 
classification~\cite{Sarinnapakorn:2007:CST:1313047.1313197}
and zero-shot classification~\cite{lampert13AwAPAMI}.
A more detailed survey of transfer learning is given by \cite{pan2009transfer_survey}.
Our method is a specific form of transfer learning, termed domain adaptation \cite{Ben-David:2010:TLD:1745449.1745461,Weinberger:2009:FHL:1553374.1553516, cross-modal_transfer}. 
Nonetheless, different from previous domain adaptation approaches, we assume the the {\em synthetic gap} is caused by the 
shift in feature distribution of synthetic data from real data and so we assume that the main 'characters' 
and 'patterns' strongly co-exist in both the synthetic and real data.
Our SMCAE is thus developed based on this assumption.

\noindent \textbf{Autoencoder} is a special type of a neural network where the
output vectors have the same dimensionality as the input vectors \cite{vincent2008ICML}. 
Autoencoder with its different variants \cite{MarginalizedDenoisingAutoencoders2012ICML,Glorot11domainadaptation, face_age_bmvc2014, face_bmvc2014}
was shown to be successful in learning and transferring shared knowledge
among data source from different domains \cite{BP:12,BY:12,DJ:13},
and thus benefit other machine learning tasks. Our framework borrows the idea of autoencoder to jointly learn two different and yet related tasks: mapping synthetic to real data; and real to real data. It is worth noting that in \cite{srivastava2012multimodal}, a multimodal autoencoder with structure similar to ours is proposed. Their multimodal autoencoder put two normal autoencoders together by sharing a hidden layer. In their structure, data at input end and output end are fully symmetric and each modal of data  occupy one branch of the antuencoder. In contrast to their structure, the proposed SMCAE composes the structure of both normal autoencoder and denoising autoencoder. With this composition, one branch of SMCAE is capable exploring intrinsic features of data in one domain, and another branch of SMCAE is going to transfer data from one domain to another domain using features discovered from both branches. The structure of SMCAE could be easily expanded to more branches to compensate more complicated multi-task learning problems. Our experiments show that our SMCAE is better than other autoencoders in this regard. 

\noindent \textbf{Learning from synthetic templates.} Some recent works of learning from synthetic data~\cite{VT:03,VT:04,BG:08} mostly generate synthetic data either by applying a simple geometric
transformation or adding image degradation to real data. To help offline recognition of handwritten text~\cite{VT:03,VT:04}, a perturbation model combined with morphological operation is applied to real data. To enhance the quality of degraded document \cite{BG:08}, degradation models such as brightness degradation, blurring degradation, noise degradation, and texture-blending degradation, were used to create a training dataset for a handwritten text recognition problem. These  methods did not address the synthetic gap problem, and thus have been limited to a small performance improvements by using synthetic data. In \cite{pishchulin2011learning}, computer graphics 3D models are used to ease training data generation. To simulate pedestrian in a picture, authors track volunteers pose from multiple views and human bodies are reshaped using a morphable 3D human model. The reshaped picture of human bodies later are composed with real world backgrounds. The same idea has been adopted in \cite{eltit} where in addition to render a 3D model to simulate an object in a real scene, features extracted from synthetic data are adapted to better train an object detector.

\section{Stacked Multichannel Autoencoder (SMCAE) }

\begin{figure}[ht]
\centerline{ %
\begin{tabular}{c}
\resizebox{0.4\textwidth}{!}{\rotatebox{0}{ 
\includegraphics{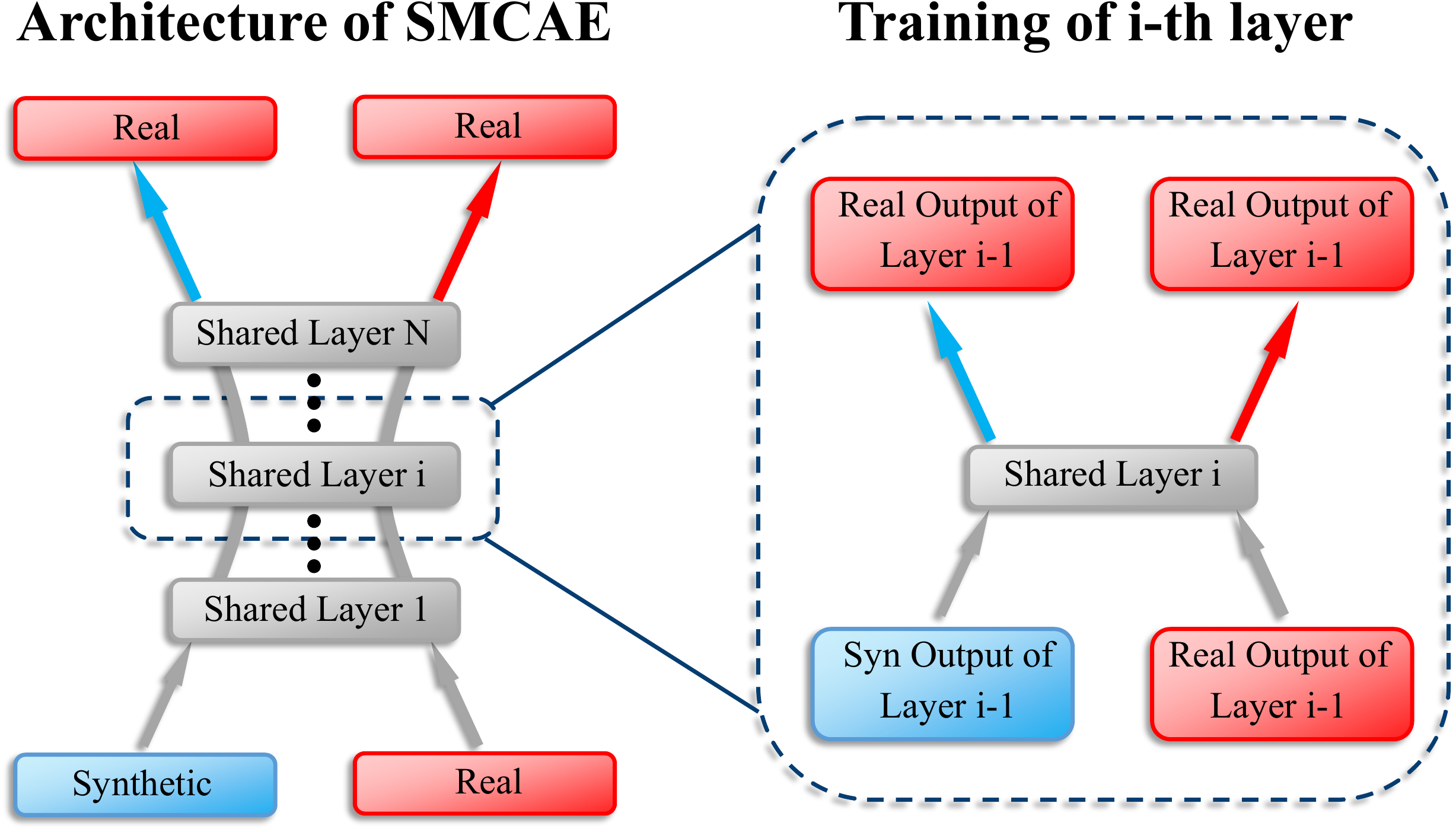}}}
\tabularnewline
\end{tabular}} \protect\protect\caption{(left) Illustration of the SMCAE: black edges between two layers are linked
to and shared by two tasks; red and blue links are separately connected
to the left and right task respectively. (right) A zoom-in structure of
SMCAE with single hidden layer. }
\label{fig: YAE-structure} 
\end{figure}

We propose the SMACE model to learn a mapping from synthetic
and real data. To learn this mapping, the SMCAE model is formulated as a stacked structure of multichannel autoencoders which facilitates an efficient and flexible way of jointly learning from both synthetic and real data. The structure and configuration of the SMCAE is illustrated in Fig. \ref{fig: YAE-structure}. Specifically, we set the left and right tasks in two channels of the SMCAE respectively. The \textit{left task}, as illustrated in left channel of Fig.~\ref{fig: YAE-structure}, takes synthetic data as input and real data as reconstruction target; while the \textit{right task} of the right channel in Fig.~\ref{fig: YAE-structure} uses real data in both input and reconstruction target. All between-layer connections that are colored in gray are shared by tasks of the two channels. The SMCAE structured in this way attempts to transform synthetic data to real data in \textit{left task} using representation learned from real data in \textit{right task}. 

\subsection{Problem setup }

We first illustarte the setup of a single layer in each channel of our SMCAE. For a single channel of our SMCAE is basically an autoencoder \cite{bengio2009learning}\cite{vincent2008extracting}. Assume an input dataset with $n$ instances $X=\{x_{i}\}_{i=1}^{n}$ where $x_{i}\in{\mathbb{R}^{m}}$. To encode the input data, we have $h_{e}(x_{i})=f(W_{e}^{j}x_{i}+b_{e}^{j})$ where $f(\cdot)$ is a sigmoid function and $\theta_{e}=\{W_{e}^{j},b_{e}^{j}\}$,
$W_{e}^j\in\mathbb{R}^{k\times m},b_{e}^j\in\mathbb{R}^{k}$ is a set
of encoding parameters in $j$-th layer. In contrast, the decoding
process is defined as $h_{d}(x_{i})=f(W_{d}^{j}h_{e}(x_{i})+b_{d}^{j})$
with the decoding parameters $\theta_{d}=\{W_{d}^{j},b_{d}^{j}\}$,
$W_{d}^j\in\mathbb{R}^{m\times k},b_{d}^j\in\mathbb{R}^{m}$ and the encoded
representations $h_{e}(x_{i})$. 

To minimize the reconstruction error, we have
\begin{equation}
\label{eq:reconstruction_error_basic}
J(\theta_{e},\theta_{d})=\frac{1}{n}\sum_{i=1}^{n}(h_{d}(x_{i})-x_{i})^{2}+\lambda W^{j}
\end{equation}
where $W^j=({\sum_k \sum_m (W_{e}^j)^{2}+\sum_m \sum_k (W_{d}^j)^{2}})/2$ is a weight decay term
added to improve generalisation of the autoencoder and $\lambda$
leverages the importance of this term. To avoid learning the identity
mapping in the autoencoder, a regularisation term $\Theta=\sum_{i=1}^{k}{\delta\text{log}\frac{\delta}{\hat{\delta}_{i}}+(1-\delta)\text{log}\frac{1-\delta}{1-\hat{\delta}_{i}}}$
that penalizes over-activation of the nodes in the hidden layer is
added%
\footnote{ $\delta$ is a sparsity parameter and is empirically set to $0.05$
in all our experiments.%
}. $\hat{\delta}_{i}$ is an averaged activation of all nodes in the hidden layer
and is computed as: $\hat{\delta}_{i}=\frac{1}{k}\sum_{i=1}^{k}h_{e}(x_{i})$.
Thus the objective of single channel is updated to:
\begin{equation}
\label{Equ: SparsityObjective}
J(\theta_{e},\theta_{d})=\frac{1}{n}\sum_{i=1}^{n}(h_{d}(x_{i})-x_{i})^{2}+\lambda W^j+\rho\Theta
\end{equation}
where $\rho$ controls sparsity of representation in hidden layer.

\subsection{The SMCAE model}

The structure of the SMCAE model is extended from an autoencoder so that it can simultaneously deal
with tasks in both the left and right channels. Specifically, we use the notation
$\langle\mathfrak{i}\text{:}X,\:\mathfrak{o}\text{:}X\rangle$ to
denote the configuration of input data (short for $\mathfrak{i}$) and
reconstruction target at the output layer (short for $\mathfrak{o}$) in one channel of
SMCAE.  We thus label the tasks in the left and right channels of SMCAE as $\langle\mathfrak{i}\text{:}X_{s},\:\mathfrak{o}\text{:}X_{r}\rangle^{L}$ and $\langle\mathfrak{i}\text{:}X_{r},\:\mathfrak{o}\text{:}X_{r}\rangle^{R}$ individually, where $\langle\cdot\rangle^{L}$ and $\langle\cdot\rangle^{R}$ indicate the left and right channel branch of SMCAE. $X_{s}$, $X_{r}$ stand for synthetic and real data respectively. The tasks in the two channels share the same parameters $\theta_{e}$ in all hidden layers which enforces the autoencoder to learn common structures of both tasks. At the output layer, we divide the SMCAE into two separate channels with their own parameters $\theta_{d}^{L}$ and $\theta_{d}^{R}$.

Our target is to minimize the reconstruction error of the two tasks of SMCAE together while taking into account the balance
between two channels. The new objective function of SMCAE is thus,
\begin{equation}
E=J^{L}(\theta_{e},\theta_{d}^{L})+J^{R}(\theta_{e},\theta_{d}^{R})+\gamma\Psi\label{Equ: YAE-objective}
\end{equation}
We add $\Psi=\frac{1}{2}(J^{L}(\theta_{e},\theta_{d}^{L})-J^{R}(\theta_{e},\theta_{d}^{R}))^{2}$
as a regularisation term to balance the learning rate between the two
channels. 

The regularization term of $\Psi$ is a novel contribution of our SMCAE. Basically, $\Psi$ penalizes a situation where the difference of learning errors between two channels are large. Since in the configuration of the SMCAE the data at the input and output end of two channels are not symmetric, the learning error resulted by optimizing learning process in two channels are very different. Having $\Psi$ in our objective will prevent from a situation where the optimization of one channel dominates the entire SMCAE so as to help SMCAE to better leverage the learning process and find a compromising balance between two channels. For importance of $\Psi$ in our objective, we show the learning results of setting different $\gamma$ for $\Psi$ in Fig. \ref{fig: rank1-parameter}

The minimization of Eq.~\ref{Equ: YAE-objective} is achieved by back propagation and stochastic gradient descent using a Quasi-Newton method -- LBFGS. In the SMCAE, with balance regularization added to the objective,
the only difference as opposed to sparse autoencoder is the gradient
computation of unknown parameters $\theta_{e}$ and $\theta_{d}^{L},\theta_{d}^{R}$.
We clarify these differences in the following equations:

\begin{equation}
\begin{split}\nabla_{W_{e}^j}E= & \frac{\partial{J^{L}}}{\partial{W_{e}^j}}+\frac{\partial{J^{R}}}{\partial{W_{e}^j}}+\gamma(J^{L}-J^{R})(\frac{\partial{J^{L}}}{\partial{W_{e}^j}}-\frac{\partial{J^{R}}}{\partial{W_{e}^j}})\\
\nabla_{b_{e}^j}E= & \frac{\partial{J^{L}}}{\partial{b_{e}^j}}+\frac{\partial{J^{R}}}{\partial{b_{e}^j}}+\gamma(J^{L}-J^{R})(\frac{\partial{J^{L}}}{\partial{b_{e}^j}}-\frac{\partial{J^{R}}}{\partial{b_{e}^j}})
\end{split}
\end{equation}
and 
\begin{equation}
\begin{split} & \nabla_{W_{d}^{L}}E=\frac{\partial{J^{L}}}{\partial{W_{d}^{L}}}+\gamma(J^{L}-J^{R})\frac{\partial{J^{L}}}{\partial{W_{d}^{L}}}\\
 & \nabla_{b_{d}^{L}}E=\frac{\partial{J^{L}}}{\partial{b_{d}^{L}}}+\gamma(J^{L}-J^{R})\frac{\partial{J^{L}}}{\partial{b_{d}^{L}}}\\
 & \nabla_{W_{d}^{R}}E=\frac{\partial{J^{R}}}{\partial{W_{d}^{R}}}+\gamma(J^{L}-J^{R})(-\frac{\partial{J^{R}}}{\partial{W_{d}^{R}}})\\
 & \nabla_{b_{d}^{R}}E=\frac{\partial{J^{R}}}{\partial{b_{d}^{R}}}+\gamma(J^{L}-J^{R})(-\frac{\partial{J^{R}}}{\partial{b_{d}^{R}}})
\end{split}
\end{equation}

We train a SMCAE in a greedy manner where one layer gets trained at a time. The configuration for training one layer of SMCAE is shown in Fig.~\ref{fig: YAE-structure}(right). The output of a trained layer is then sent as input to the next layer for training. A fine-tuning is implemented to the entire stacked structure once all layers are trained. Thus,  after SMCAE has been trained, to transform new synthetic data, the data is sent to the left channel of the SMCAE $\langle\mathfrak{i}\text{:}X_{s},\:\mathfrak{o}\text{:}X_{r}\rangle^{L}$. We take output of this process as transformed synthetic data.

\subsection{Competitors}
\begin{figure}[ht]
\centerline{ %
\begin{tabular}{cccc}
\resizebox{0.12\textwidth}{!}{\rotatebox{0}{ 
\includegraphics{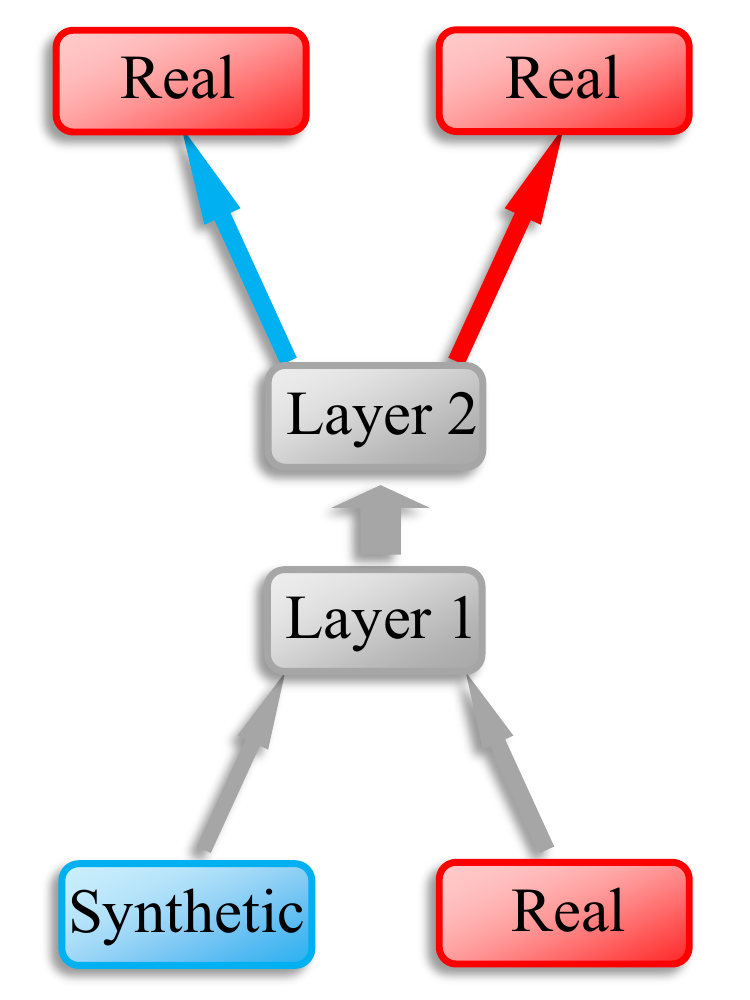}}}
&
\resizebox{0.12\textwidth}{!}{\rotatebox{0}{ 
\includegraphics{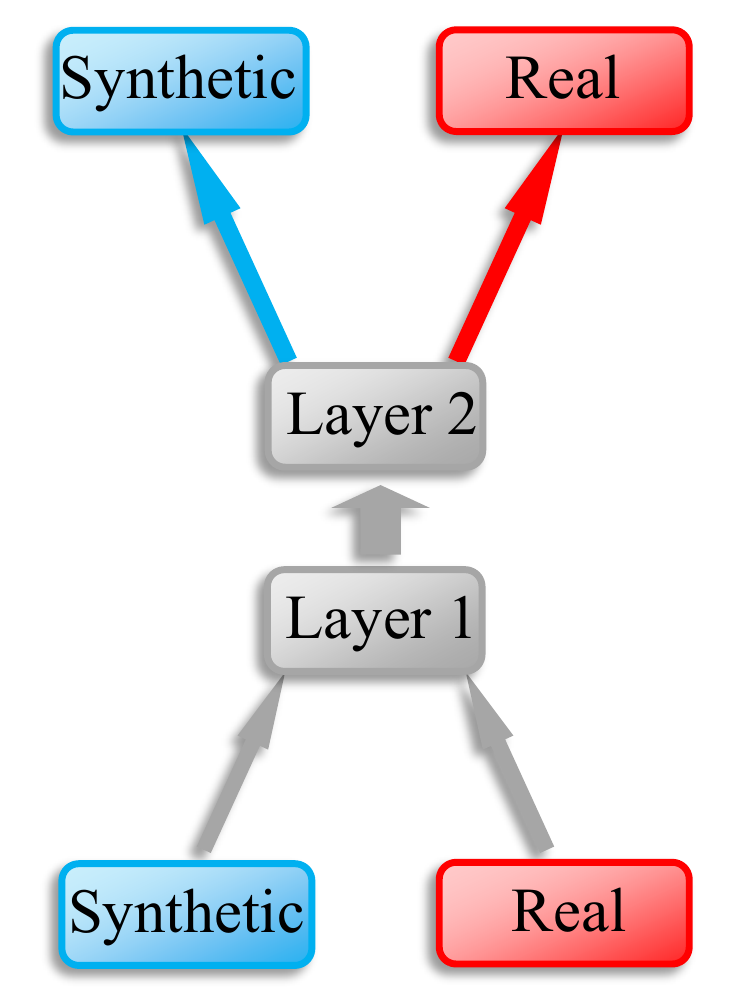}}}
&
\resizebox{0.12\textwidth}{!}{\rotatebox{0}{ 
\includegraphics{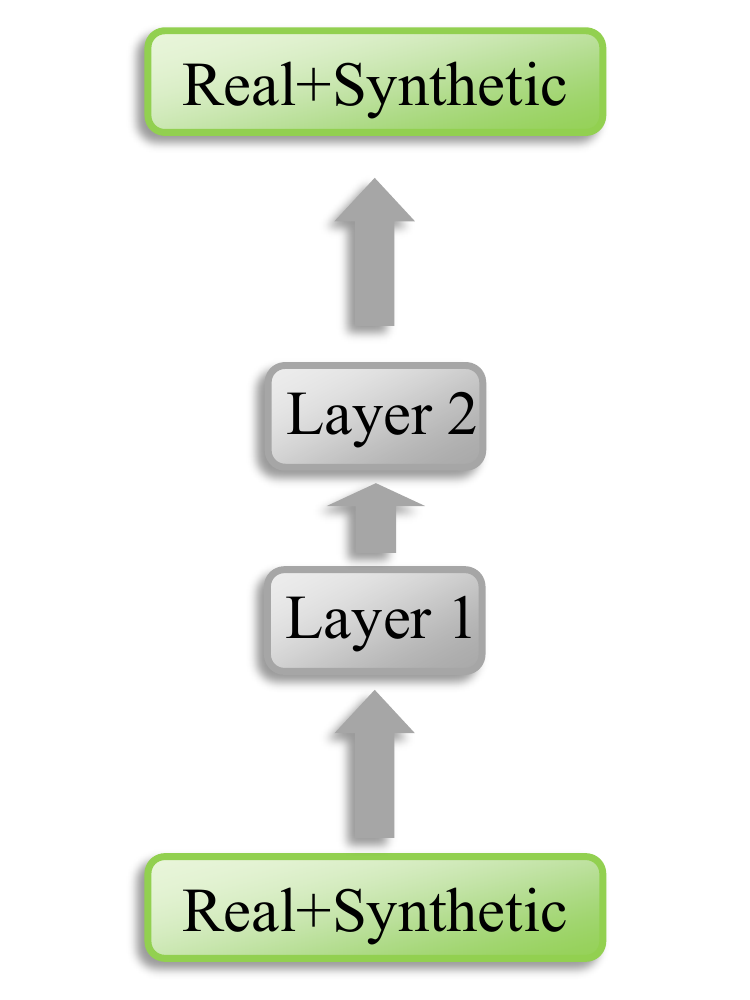}}}
&
\resizebox{0.12\textwidth}{!}{\rotatebox{0}{ 
\includegraphics{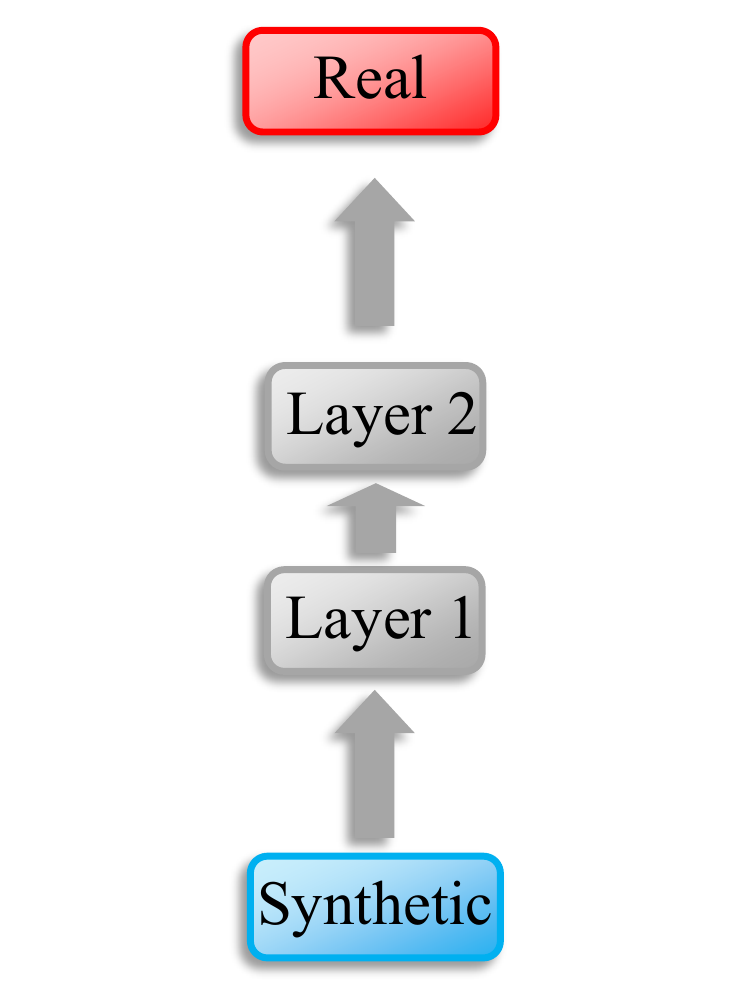}}}
\\
SMCAE & SMCAE-II & SAE-I & SAE-II
\end{tabular}} 
\caption{Illustration of the compared configurations: SMCAE, SMCAE-II, SAE-I and SAE-II.}
\label{fig: smcaecompetitors} 
\end{figure}

As shown in Fig.~\ref{fig: smcaecompetitors}, we compare the SMCAE configuration to three alternative configurations: (1) SMCAE-II which places two separate channels on the structure,
i.e. $\langle\mathfrak{i}\text{:}X_{s},\:\mathfrak{o}\text{:}X_{s}\rangle^{L}$
and $\langle\mathfrak{i}\text{:}X_{r},\:\mathfrak{o}\text{:}X_{r}\rangle^{R}$.
(2) Stacked autoencoder type-I (SAE-I) which merges the tasks in a single channel
stacked autoencoder, with the configuration of :$\langle\mathfrak{i}\text{:}X_{s}X_{r},\:\mathfrak{o}\text{:}X_{r}X_{r}\rangle$. 
(3) Stacked autoencoder type-II (SAE-II) which simply transforms source data to target data, and configures as:
$\langle\mathfrak{i}\text{:}X_{s},\:\mathfrak{o}\text{:}X_{r}\rangle$. 

 Compared with  SAE-I and SAE-II, our two channel structures endow more flexibility. Critically, the single channel models force  synthetic data to fit real data, which causes synthetic data to lose information and become less useful for recognition. In contrast, SMCAE can explore `characters' and `patterns' common in  both synthetic and real data. Intrinsically, SMCAE first encodes both synthetic and real data into common hidden layers which model common information useful for recognition. Then the decoding process transforms the synthetic data to better simulate real data. Although SMCAE-II has the same two branches in the structure, it does not learn such transformation between synthetic data and real data.

\section{Experiments and Results}
 We first compare  SMCAE on the challenging task of face-sketch recognition \cite{wang2009face,zhang2011coupled} using the CUFSF dataset. We show that  SMCAE is better than alternative configurations. To further validate the efficacy of our framework,  we train SMCAE on handwritten digit images and generate synthetic data to simulate real images. We show that the synthetic data can help train  classifiers for recognition. 

\noindent \textbf{\small{}Dataset.} We conduct our experiments on two different
datasets: (1) The CUFSF dataset~\cite{wang2009face,zhang2011coupled} containing the photos and sketches of $1194$ people with lighting variations. We employ the standard split defined in~\cite{wang2009face,zhang2011coupled}
which selects $500$ persons as the training set, and the remaining
$694$ persons as the testing set. (2) handwritten digits dataset%
\footnote{collected from UCI machine learning repository (HWDUCI)~\cite{Bache+Lichman:2013}.
} (HWDUCI) containing $5620$ instances in total in which $3823$ samples
are used for training and $1917$ samples are used for testing. The
handwritten digits from $0$ to $9$ in this dataset are collected
from $43$ people: $30$ contributed to the training set and the other
$13$ to the test set. For all experiments, we empirically set the number of hidden layers in SMCAE to two and each layer has 1000 nodes. The same settings are used to make SMCAE, SMCAE-II, SAE-I and SAE-II more comparable.

\noindent \textbf{\small{}Evaluation Metrics. }We report the following
metrics when they are available: (1) F1-score, which is defined as
$F1=2\cdot\left(Precision\cdot Recall\right)/\left(Precision+Recall\right)$.
(2)Receiving Operator Characteristic (ROC) curves and VR@0.1$\%$FAR
which is the performance of Verification Rate (VR) at 0.1$\%$ False
Acceptance Rate (FAR). VR@0.1$\%$FAR is a standard evaluation metric
and proposed in \cite{wang2009face}. (3) Rank-1 recognition accuracy.

\noindent \textbf{\small{}Features.}(1) Similar to \cite{klare2011matching}, in the CUFSF
dataset we use Histogram of Oriented Gradients (HOG).   To further reduce the computational cost, the resolution
of all photos and sketches is reduced to $50\times50$. So the cell
size of HOG features is set to 3. 
(2)The HWDUCI dataset uses HOG features with cell size 3. 

\noindent \textbf{\small{}Classifiers.} For CUFSF dataset, nearest-neighbor search with Euclidean metric is used in retrieving the most similar photo to the query sketch. In the handwritten digit classification, a Support Vector Machine (SVM) with RBF kernel\footnote{The parameters are cross-validated} is used in the experiments.

\subsection{Results on the CUFSF dataset}
\label{subsec: results cufsf}

In all experiments on this dataset, HOG features of sketch images are first transformed by the SMCAE and then used as queries. We first compare the results of photo-sketch matching using HOG feature transformed by SMCAE, SMCAE-II, SAE-I and SAE-II. The results are reported as ROC curve starting with VR@0.1$\%$FAR. The dissimilarity between a photo and a sketch is computed as the Euclidean distance between descriptors. 

\begin{figure}[ht]
\centerline{ %
\begin{tabular}{c}
\resizebox{0.45\textwidth}{!}{{ \includegraphics{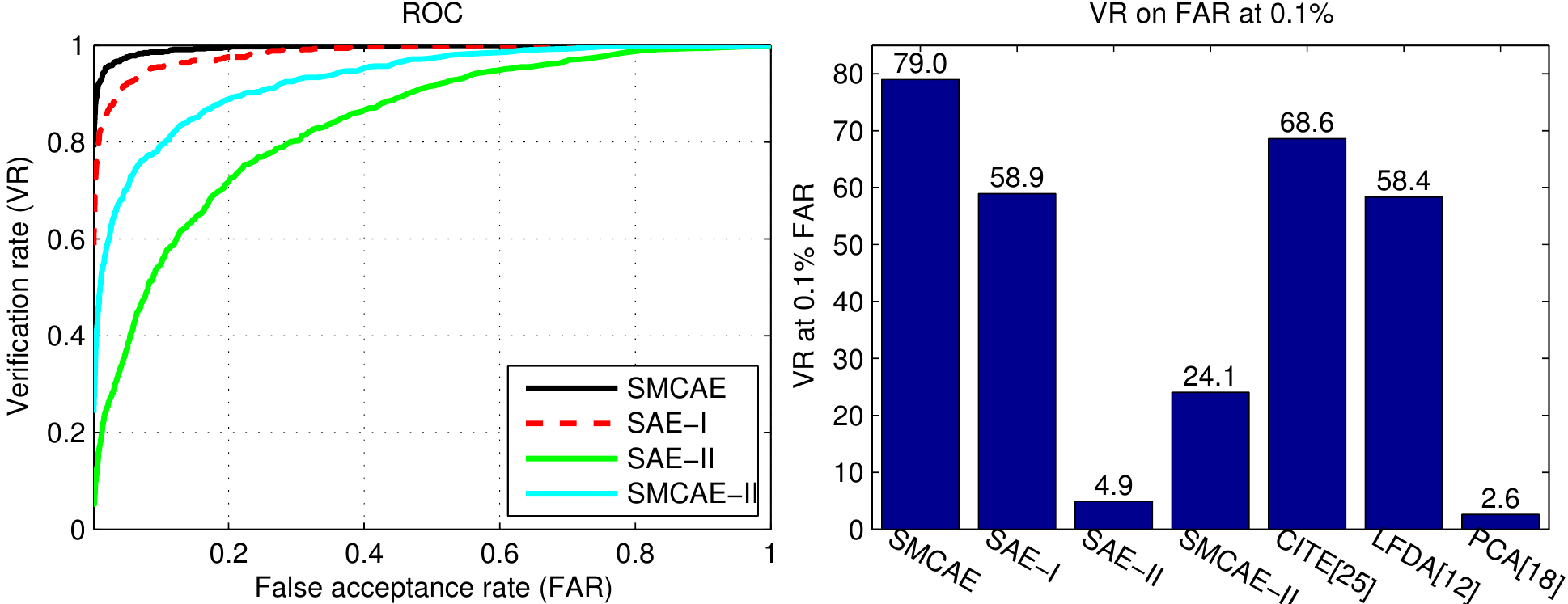}}}  
\end{tabular}} 
\caption{Results on CUFSF dataset. Left: ROC curve of different methods; Right: VR@0.1$\%$FAR of different methods.}
\label{fig: feretrocs}
\end{figure}

The ROC curves and   VR@0.1$\%$FAR are shown in Fig. \ref{fig: feretrocs}. Clearly, the proposed SMCAE achieves the highest results on AUC values and VR@0.1$\%$FAR accuracy and significantly outperforms the alternative configurations. Note that we also report  the state-of-the-art approaches   of VR@0.1$\%$FAR including LFDA \cite{klare2011matching}, CITE \cite{zhang2011coupled} and classic eigenfaces(PCA)\cite{turk1991face}. It is worth noting that in some of previous works, a better result could be obtained by combining multiple features. For example in \cite{zhang2011coupled}, multiple CITE features generated by a random forest are used to batter matching photos and sketches. Here, to enable a comparison with more fairness, we focus our comparison on matching results obtained by using uncombined feature only.

\begin{figure}[ht]
\centerline{ %
\begin{tabular}{c}
\resizebox{0.33\textwidth}{!}{{ \includegraphics{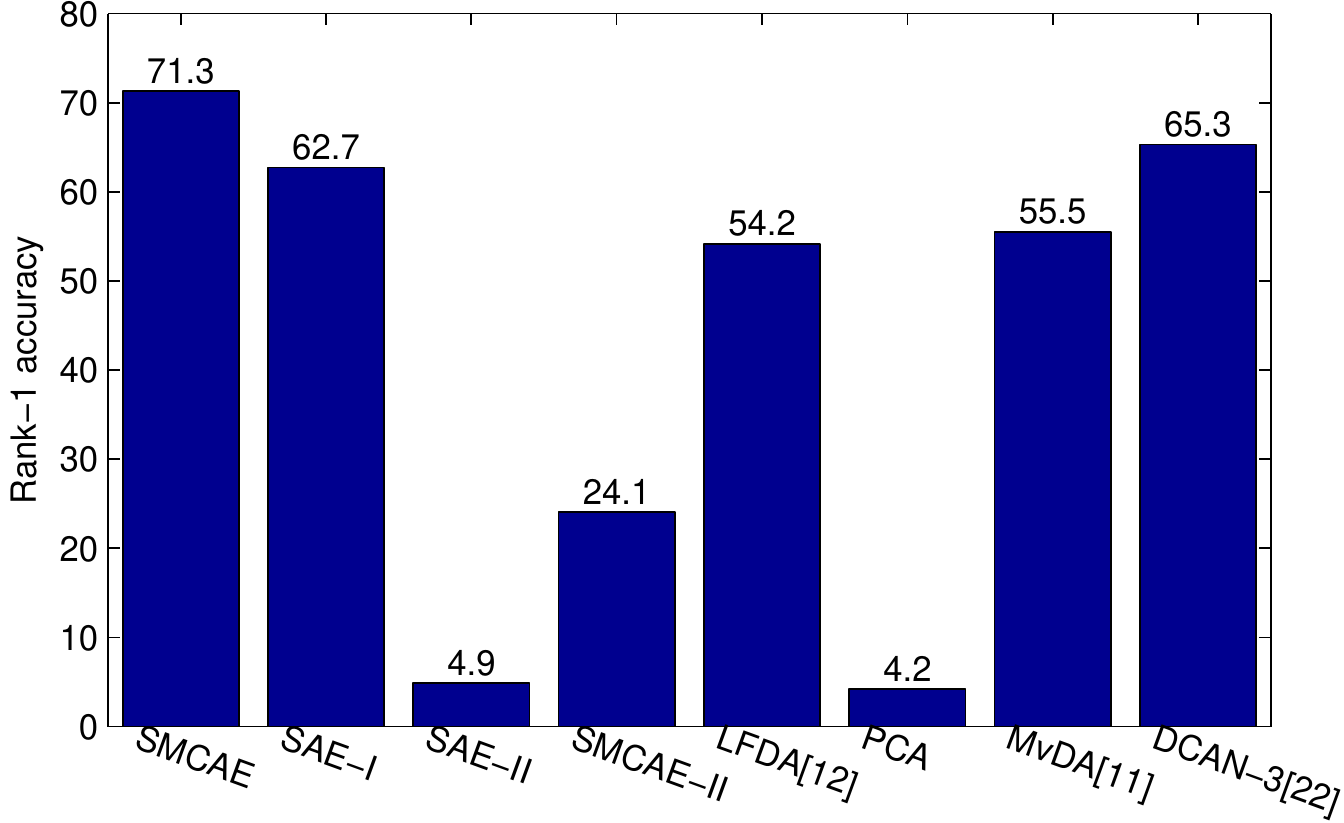}}}  
\end{tabular}} 
\caption{Rank-1 accuracy of different methods on CUFSF dataset.}
\label{fig: rank1}
\end{figure}

There are several reasons why our SMCAE outperform the other approaches. First, compared with SMCAE-II, the configuration of SMCAE involves a task that handles the transformation from synthetic to real data, and thus better eliminates the distance between them. Second, compared with SAE-I, rather than  merging two tasks in a single channel  SMCAE  employs  two channels to better clarify each task with the aim of reconstructing the main `characters' and `patterns' co-existing in both tasks. Thus synthetic data can be more easily transformed to real data with less error. Finally, SMCAE is better than SAE-II as SMCAE learns features of real data in task $\langle\mathfrak{i}\text{:}X_{r},\:\mathfrak{o}\text{:}X_{r}\rangle^{R}$. These features will better compensate the difference between synthetic data and real data during the transformation.

\begin{figure}[ht]
\centerline{ %
\begin{tabular}{c}
\resizebox{0.4\textwidth}{!}{{ \includegraphics{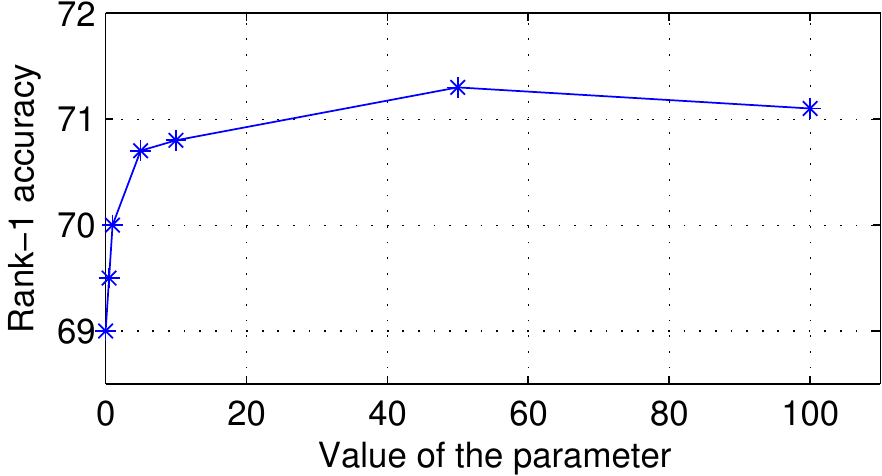}}}  
\end{tabular}} 
\caption{Rank-1  accuracy by setting different value for $\gamma$ in Eq~ \ref{Equ: YAE-objective}. Rank-1 accuracy by setting $\gamma$ equal to 0, 0.5, 1, 5, 10, 50, and 100 are shown in the figure.}
\label{fig: rank1-parameter}
\end{figure}

We further validate the results by using  Rank-1 recognition accuracy which is also reported in \cite{kan2012multi,wang2014deeply}. The results are shown in Fig.~\ref{fig: rank1}. The methods of \cite{kan2012multi,wang2014deeply} are comparable to our SMCAE.  Method \cite{kan2012multi} employed a discriminant common subspace to maximize the between-class variations and minimize the within-class variations. Method~\cite{wang2014deeply}  used a structure composed of two autoencoders. As can be seen Fig.~\ref{fig: rank1}, the SMCAE outperforms all other methods.

\noindent \textbf{\small{ Parameter Validation  in Eq.~\ref{Equ: YAE-objective}}}.  To validate the significance of $\Psi$ in Eq.~\ref{Equ: YAE-objective}. We set $\gamma$ with different values and report the rank-1  accuracy in Fig~\ref{fig: rank1-parameter}.  Particularly, when $\gamma$ is $0$, it takes 2 times longer for SMCAE to converge compared with $\gamma=50$  used in this work, Further with $\gamma=0$ the rank-1 accuracy  is dropped by more than $2\%$. This validates the importance of term $\Psi$ discussed in Sec. 3.2.

\begin{figure}[ht]
\centerline{ %
\begin{tabular}{ccccc}
\resizebox{0.08\textwidth}{!}{{ 
\includegraphics{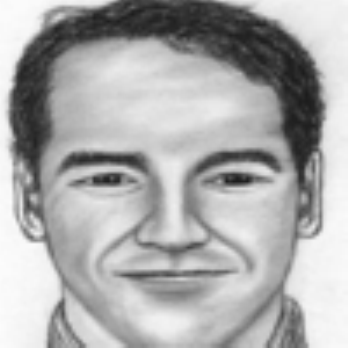}}}  & \resizebox{0.08\textwidth}{!}{{ \includegraphics{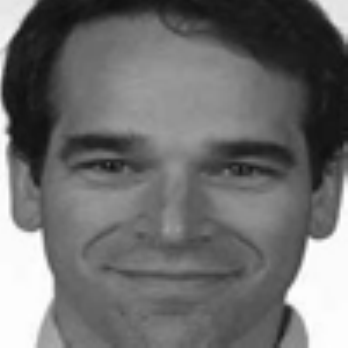}}}  & \resizebox{0.08\textwidth}{!}{{ \includegraphics{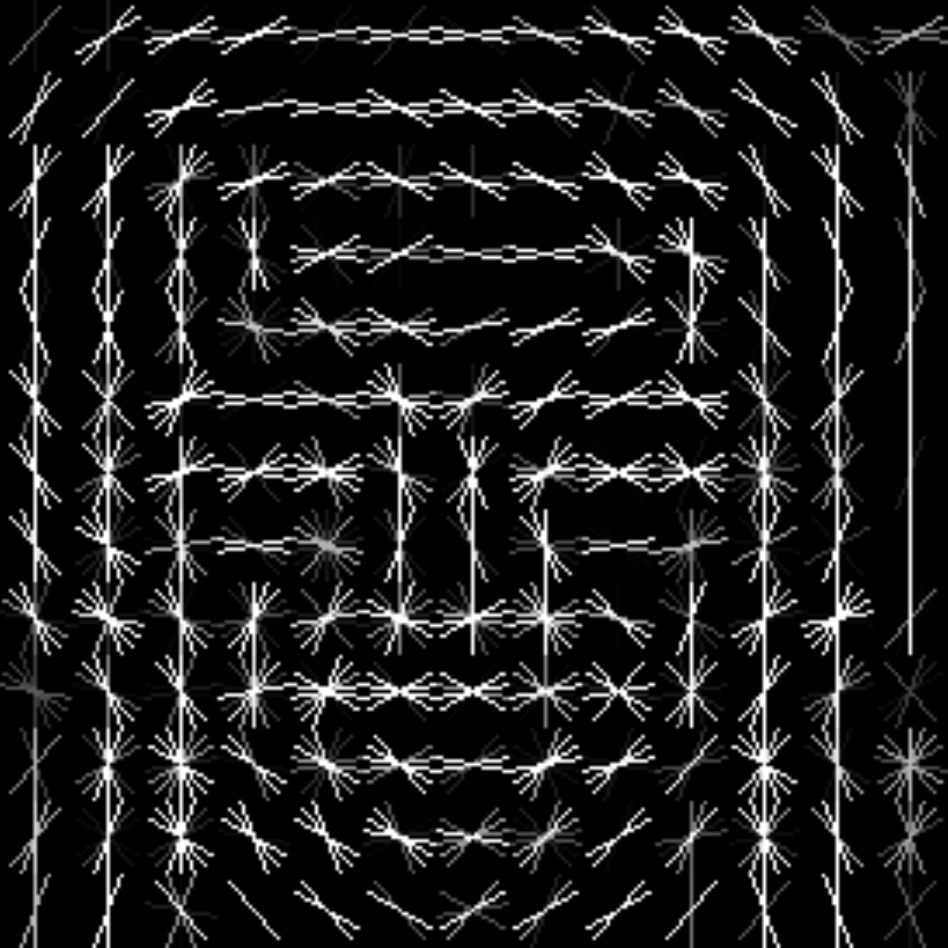}}}  & \resizebox{0.08\textwidth}{!}{{ \includegraphics{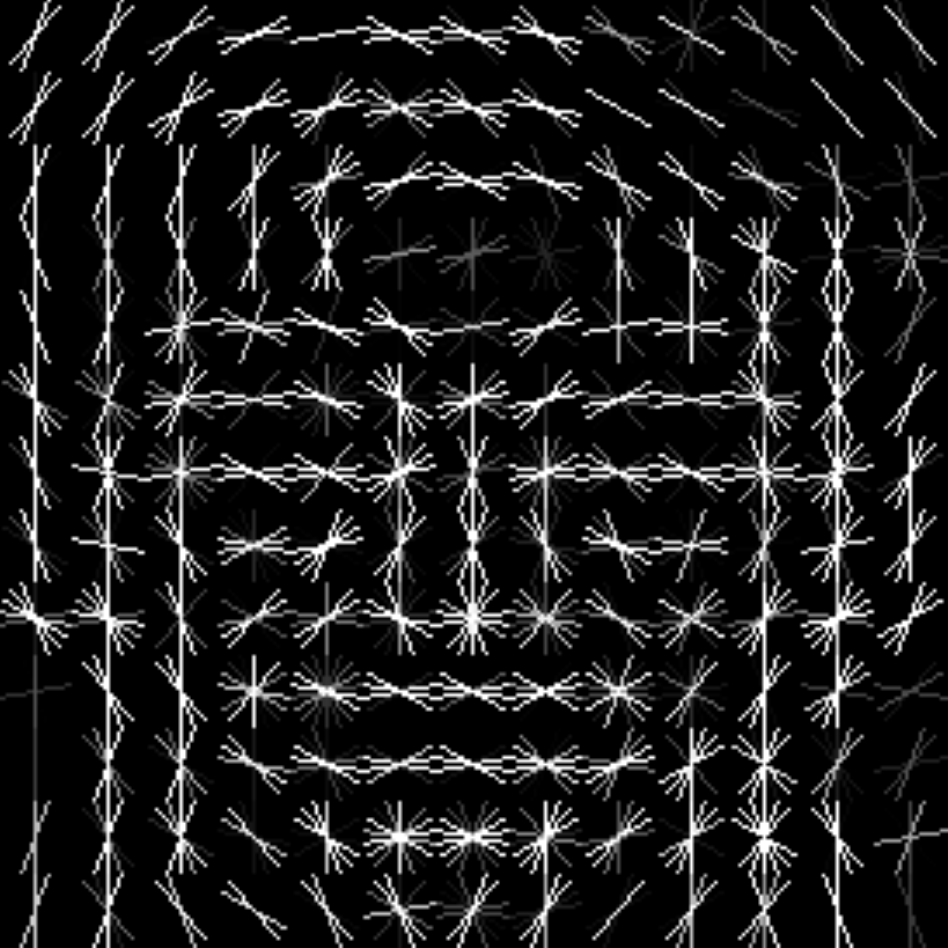}}}  & \resizebox{0.08\textwidth}{!}{{ \includegraphics{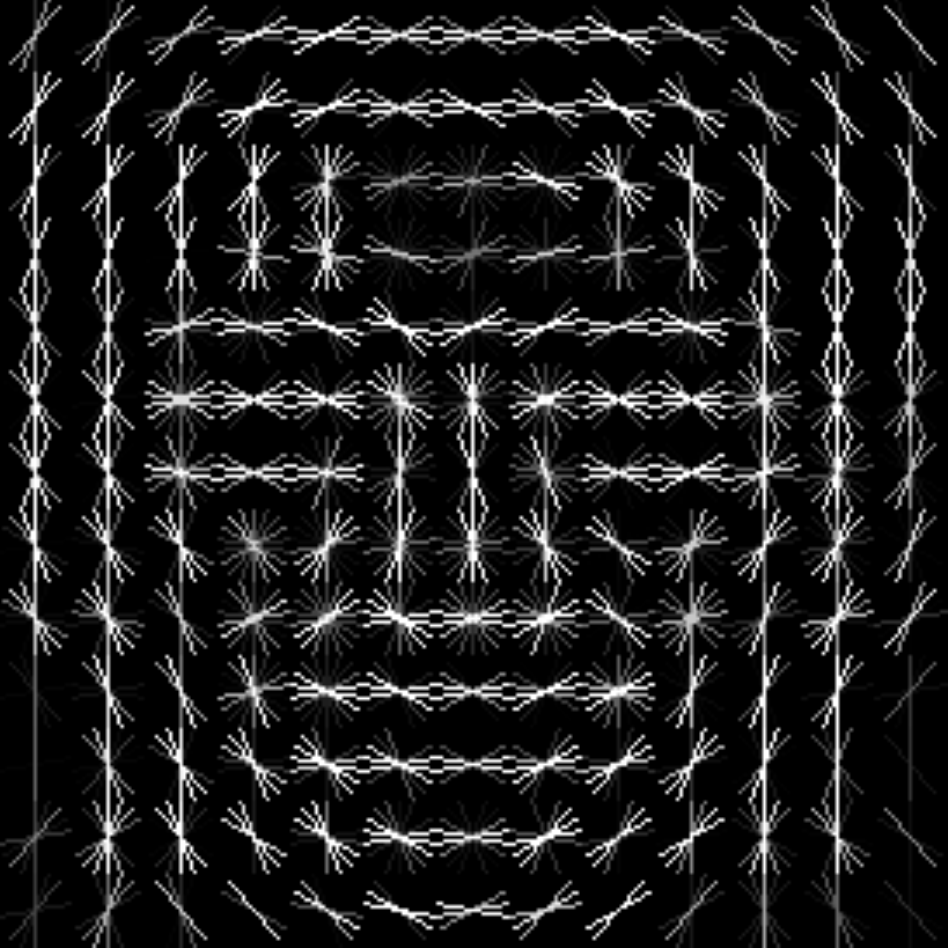}}} \tabularnewline
\resizebox{0.07\textwidth}{!}{{ 
\includegraphics{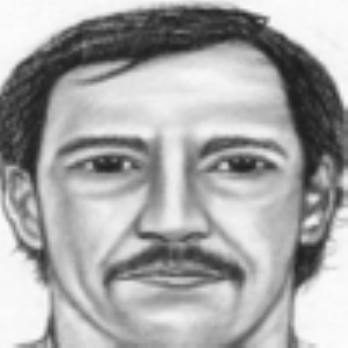}}}  & \resizebox{0.08\textwidth}{!}{{ \includegraphics{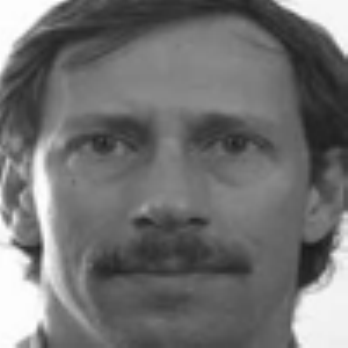}}}  & \resizebox{0.08\textwidth}{!}{{ \includegraphics{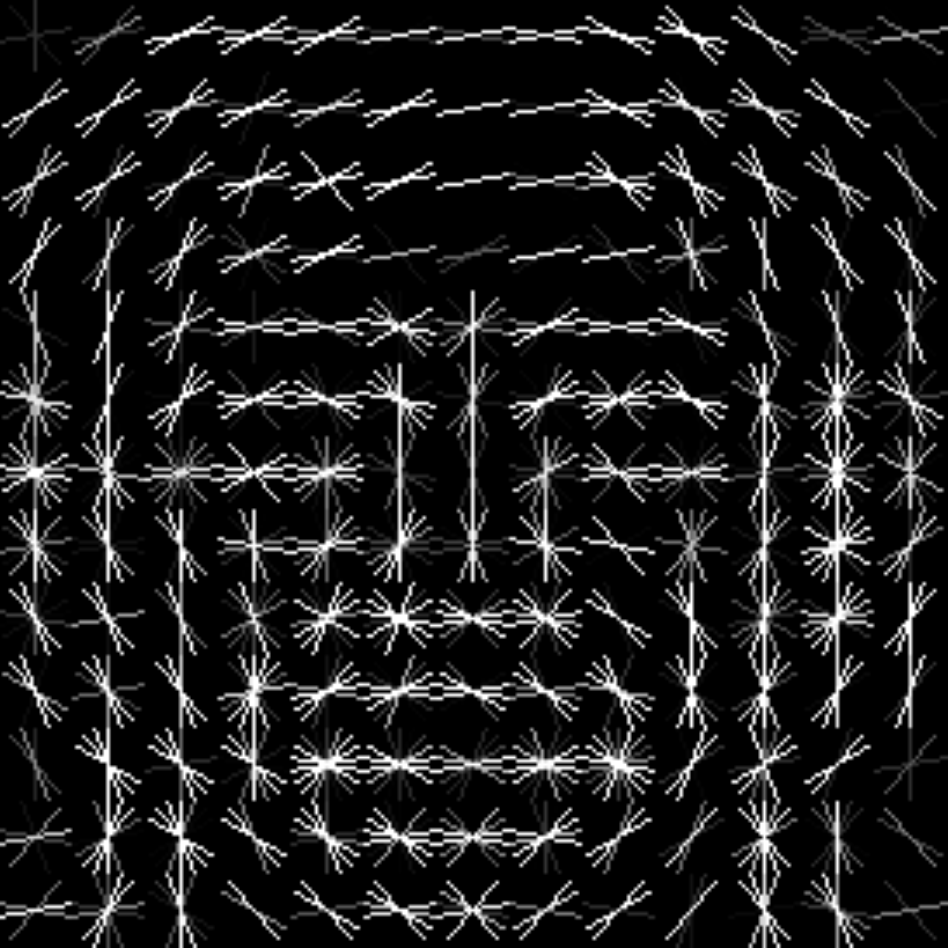}}}  & \resizebox{0.08\textwidth}{!}{{ \includegraphics{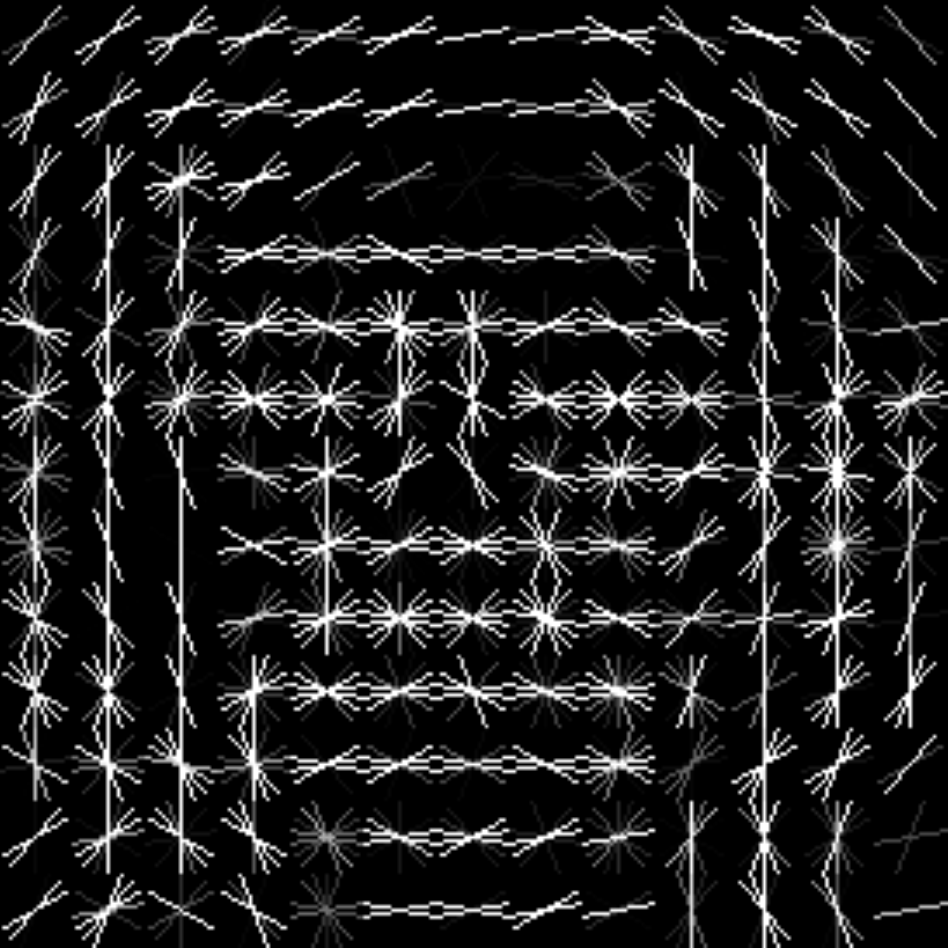}}}  & \resizebox{0.08\textwidth}{!}{{ \includegraphics{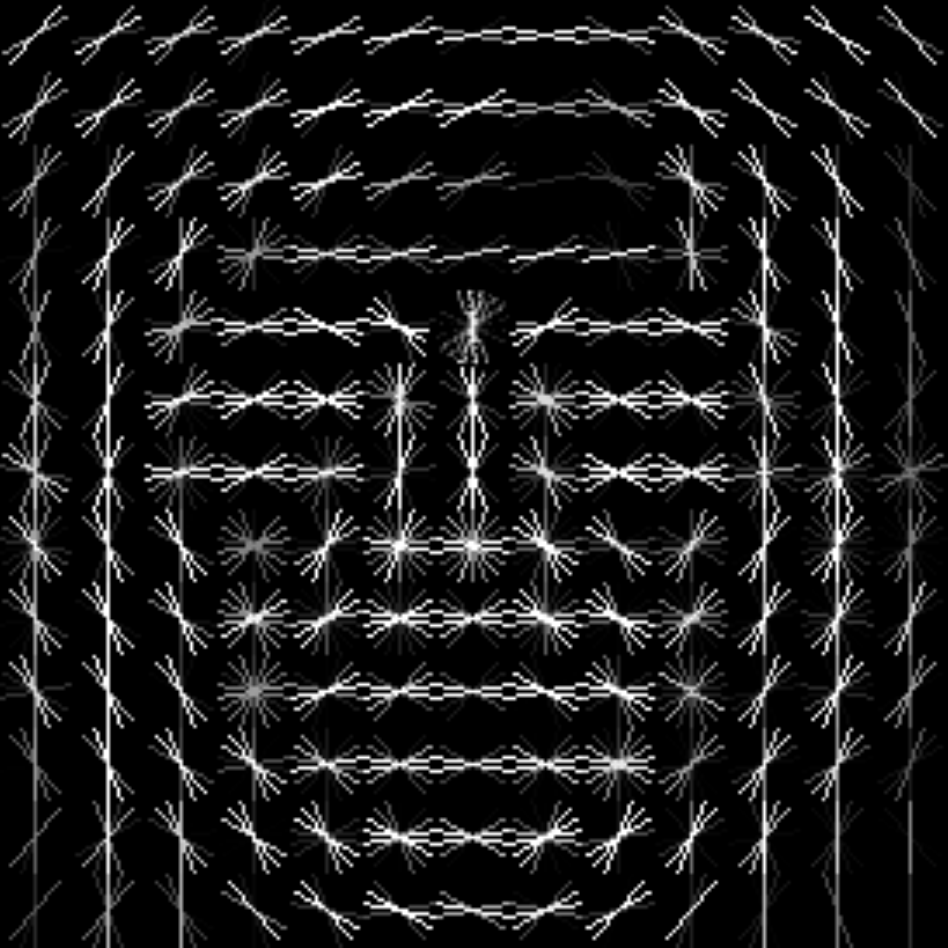}}} \tabularnewline
\resizebox{0.08\textwidth}{!}{{ 
\includegraphics{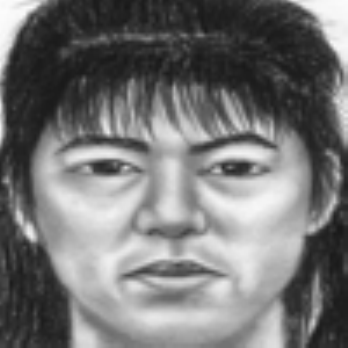}}}  & \resizebox{0.08\textwidth}{!}{{ \includegraphics{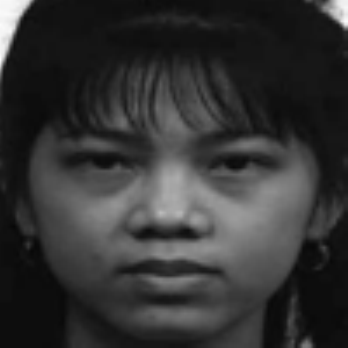}}}  & \resizebox{0.08\textwidth}{!}{{ \includegraphics{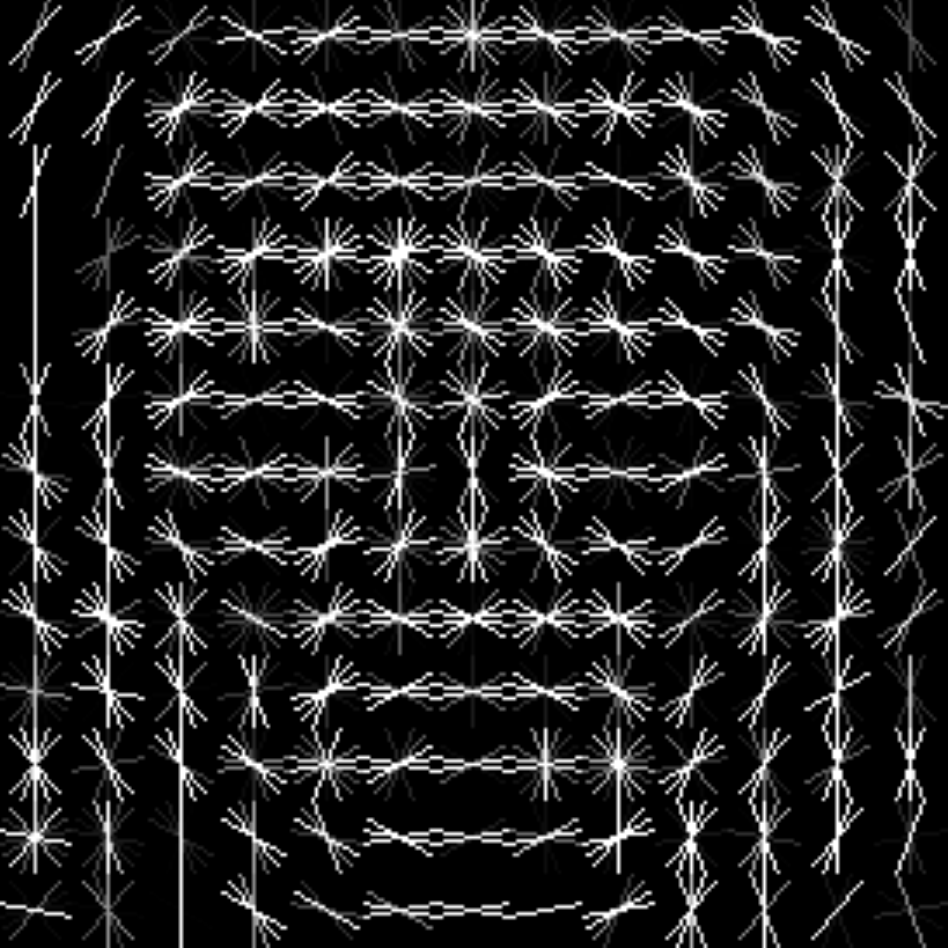}}}  & \resizebox{0.08\textwidth}{!}{{ \includegraphics{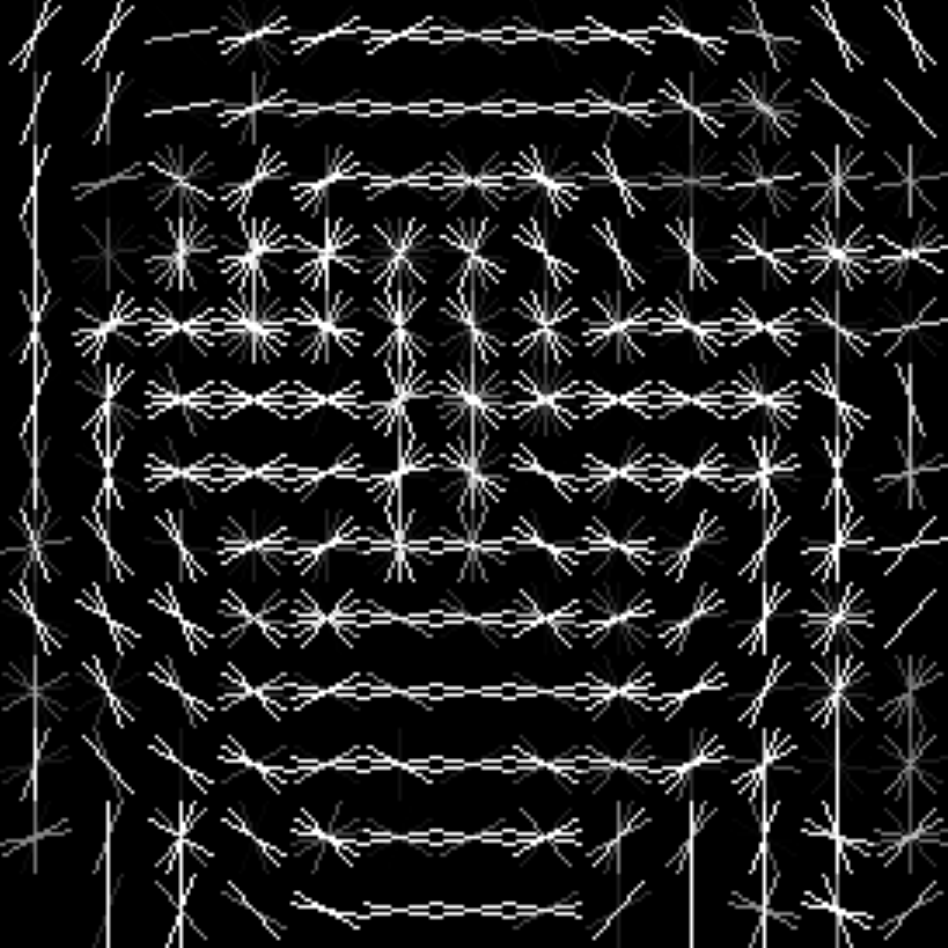}}}  & \resizebox{0.08\textwidth}{!}{{ \includegraphics{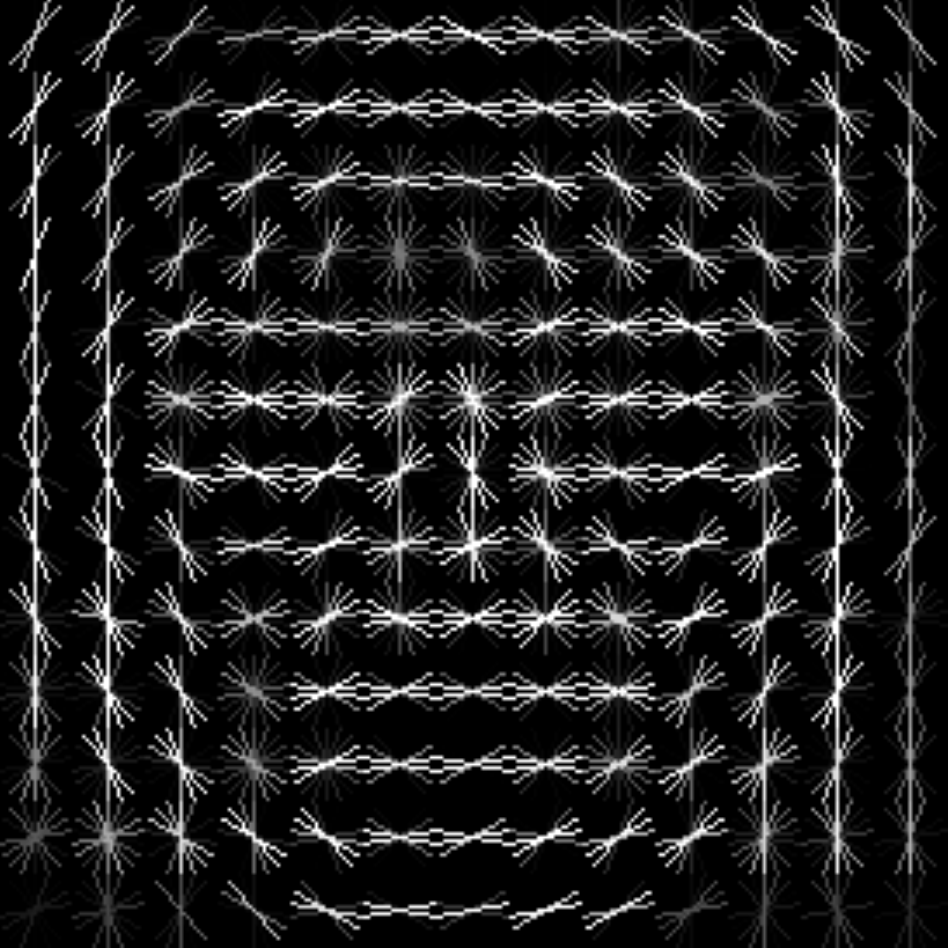}}} \tabularnewline
A  & B  & C  & D  & E \tabularnewline
\end{tabular}} 
\caption{Example of HOG features transformed by SMCAE.
A: Sketch. B: Photo. C: Original sketch HOG. D: Photo HOG. E: Transformed
sketch HOG.}
\label{fig: ferettsne}
\end{figure}

\noindent \textbf{\small{Qualitative results}}. Some qualitative
results are shown in Fig. \ref{fig: ferettsne}. It shows that a sketch
HOG transformed by our SMCAE is more similar to the ground truth photo HOG.

\subsection{Handwritten Digit Recognition}

\begin{figure}[h]
\centerline{ %
\begin{tabular}{cc}
\resizebox{0.49\textwidth}{!}{\rotatebox{0}{ \includegraphics{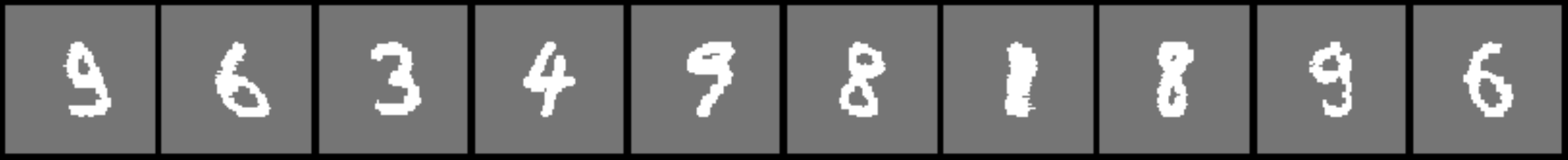}}}  & \tabularnewline
\resizebox{0.49\textwidth}{!}{\rotatebox{0}{ \includegraphics{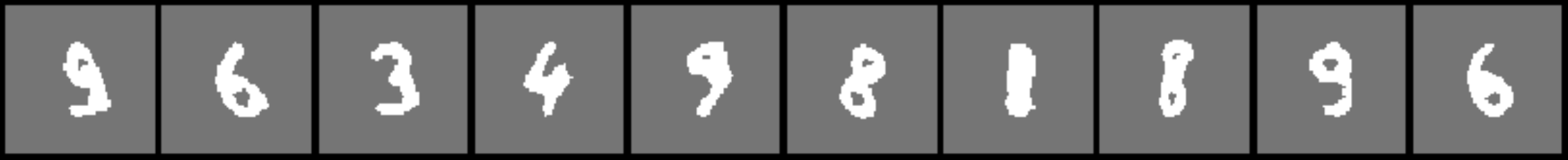}}}  & \tabularnewline
\end{tabular}} \protect\protect\caption{Illustration of real digit images (upper row) and corresponding synthetic
versions (lower row).}
\label{fig: syndigit} 
\end{figure}

\noindent \textbf{\small{Generating synthetic data}}. A synthetic version of each real character is generated as a variant of a centralized model learned from real characters. The centralized model of digit is shaped by control points $C=\{c_{i}\}_{i=1}^{n}$ settled on the boundary of the digit. A technique called migration is used to locate corresponding control points on each real digit image. A synthetic digit image then could be generated by filling areas closed by the control points \footnote{Please refer to supplementary material for details.}. Examples of generated synthetic digits are shown in Fig.~\ref{fig: syndigit}. To generate more synthetic data which is used to train the classifier once transformed by the trained SMCAE, we assume that locations of the control points follow a multivariate normal distribution $C\sim N(\mu,\Sigma)$ with $\mu$ and $\Sigma$ estimated using control points on the synthetic digit images. For each digit, $3,000$ new synthetic images are generated by randomly drawing samples from $N(\mu,\Sigma)$.

%In this experiment,
%we adopt an approach described in {[}XXX{]} to get the centralized
%model for each digit and generate synthetic version for each image.

%To this end, the real digit images and the corresponding synthetic
%images could be used to train SMCAE using the proposed configuration.
%The HoG features with cell size equal to 3 are used in the training.
%Then, to generate a lot more synthetic data for training classifier,
%we assume locations of the control points follow a multivariant normal
%distribution $C\sim N(\mu,\Sigma)$ with $\mu$ and $\Sigma$ estimated
%using control points on the synthetic digit images. For each digit,
%$3,000$ new synthetic images are generated by randomly draw samples
%from $N(\mu,\Sigma)$. The trained SMCAE then is used to transform
%total $30,000$ synthetic samples. We choose support vector machine
%trained by transformed $30,000$ samples as the classifier in our
%test.

We compare our SMCAE with SMCAE-II, SAE-I, SAE-II, LeNet-5 \cite{LY:98} and the best results \cite{Alimoglu97combiningmultiple} reported on this data set. The classification performance is evaluated by F1-score.  A Support Vector Machine (SVM) classifier with RBF kernel is used in the experiments. For SMCAE, SMCAE-II, SAE-I and SAE-II in the test, real training data together with transformed synthetic data are used to train the SVM.

\begin{figure}[ht]
\begin{centering}
\includegraphics[scale=0.48]{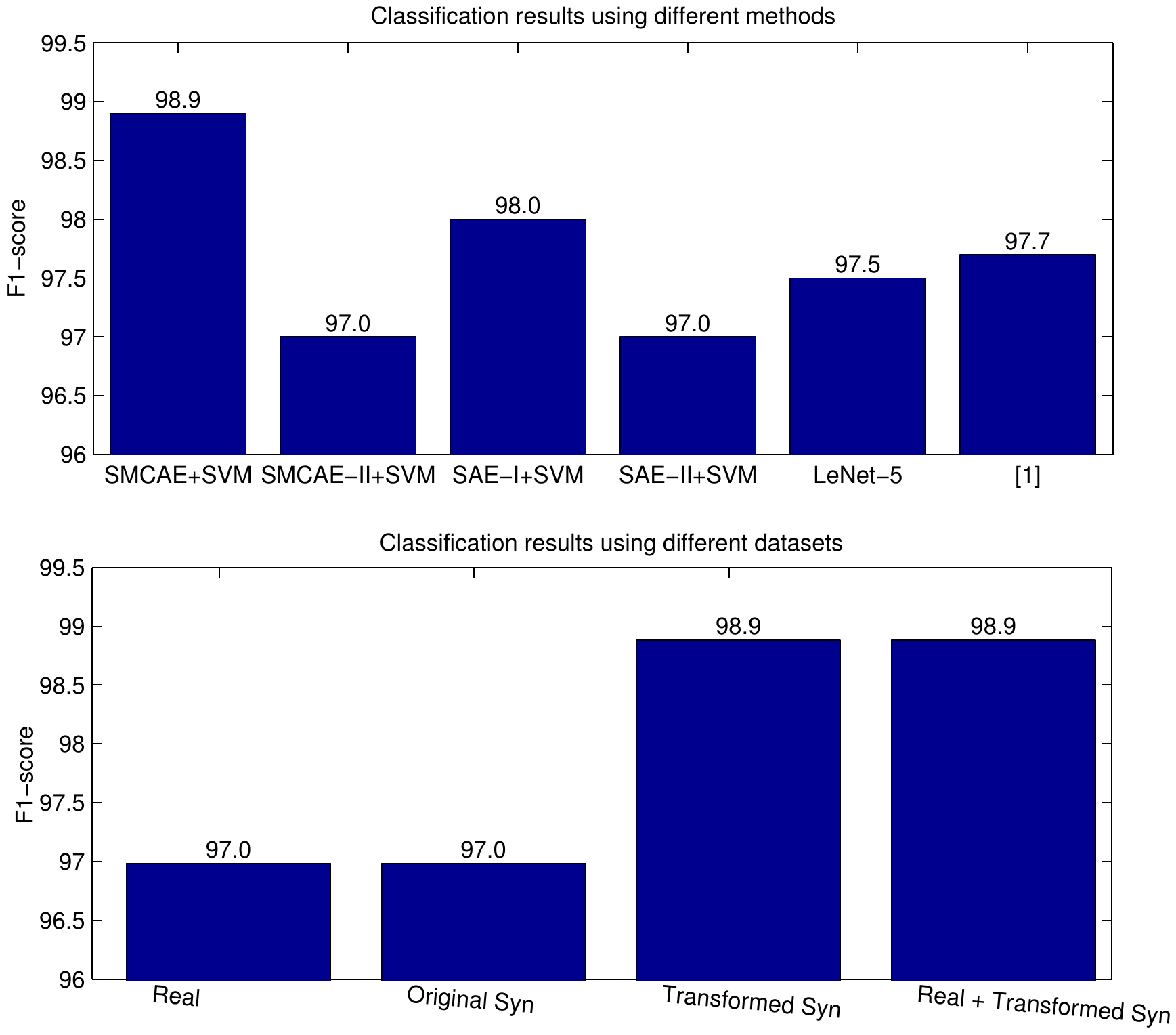}
\par\end{centering}
\protect\caption{\label{fig: digitdifferentdataset} Comparison of classification results
in F1-score of different methods (left) and different training datasets
(right). In the left figure, all the methods in the test used the same
training dataset which combines original real data and transformed
synthetic data. }
\end{figure}

\begin{figure}[ht]
\begin{centering}
\includegraphics[scale=0.65]{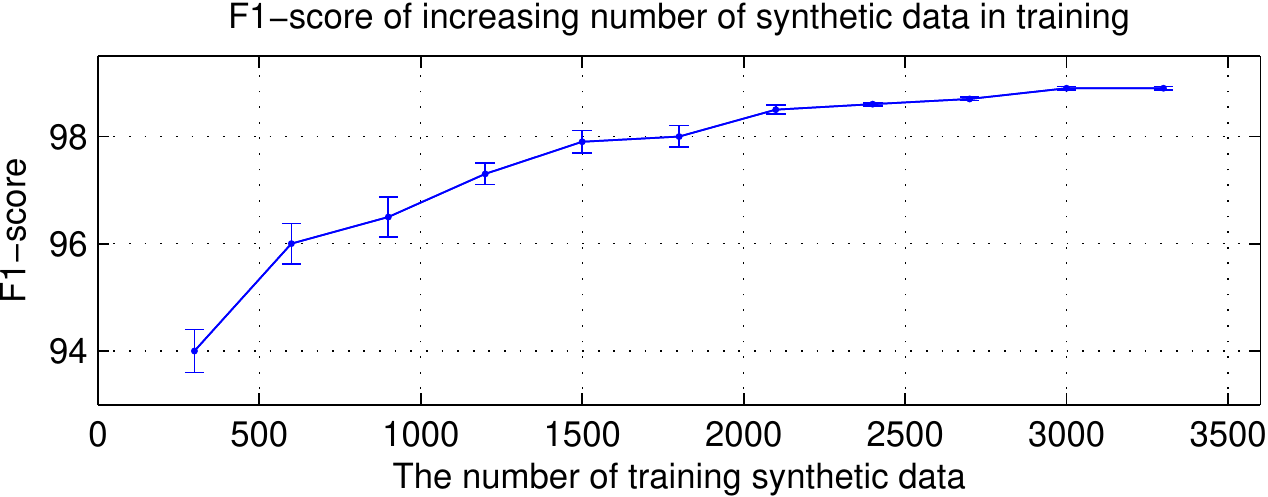}
\par\end{centering}
\protect\caption{\label{fig: increasingtrainingsyn} F1-scores and corresponding standard deviation when increasing the number of synthetic data used in training is shown.}
\end{figure}

As shown in Fig. \ref{fig: digitdifferentdataset} (left),  the SVM classifier with our SMCAE  is better than all the alternative methods. This validates the effectiveness of our framework in generating synthetic data to better help training a classifier.

To further demonstrate how transformed synthetic data improve the classification results, we conducted more evaluations by training classifiers using different combinations of training sets
in Fig.~\ref{fig: digitdifferentdataset} (right). Particularly, four combinations of training sets are used. First, to have a performance baseline of SVM, we trained the SVM using real data only. To investigate how much improvement we could obtain in classification using a SVM trained by transformed synthetic data, we compare a SVM trained by synthetic data and transformed synthetic data respectively. The best performance is obtained with a SVM trained by real data together with transformed synthetic data.

With more synthetic training data generated by SMCAE, we gain a large margin of improvement in the classification. We notice that we can get the same result ($0.989$) by using Transformed synthetic and Real+Transformed Synthetic separately in Fig.~\ref{fig: digitdifferentdataset} (right), which highlights the effectiveness of SMCAE in transforming synthetic data to simulate real data.

Finally, it is interesting to evaluate how the amount of synthetic data affects the classification results. We increasingly add more transformed synthetic data (from 300 to 3,300 samples) when training the SVM. The classification results are reported in Fig. \ref{fig: increasingtrainingsyn}. The curve shows an ascending trend when adding more samples, which means that all transformed synthetic data added to this test are highly effective and useful in the classification.

\section{Conclusion}

In this paper we identify the synthetic gap problem. To solve this problem,  we propose a novel Stacked Multichannel
autoencoder (SMCAE) model. SMCAE has multiple channels in its structure
and is an extension of a standard autoencoder. We show that SMCAE not
only bridges the synthetic gap between real data and synthetic data,
but also jointly learns from both real and synthetic data. 

{\small
\bibliographystyle{plain}
\bibliography{WithSM}
}

\clearpage

\centering{\textbf{Supplemental Material} }

\section{Optimization of SMCAE}

With two branches in the SMCAE, we target to minimize the reconstruction
error of two tasks together while taking into account the balance
between two branches. The new objective function is given as:

\begin{equation}
E=J^{L}(\theta_{e},\theta_{d}^{L})+J^{R}(\theta_{e},\theta_{d}^{R})+\gamma\Psi\label{Equ: YAE-objective}
\end{equation}
where 
\begin{equation}
\Psi=\frac{1}{2}(J^{L}(\theta_{e},\theta_{d}^{L})-J^{R}(\theta_{e},\theta_{d}^{R}))^{2}
\end{equation}
is a regularization added to balance the learning rate between two
branches. In the SMCAE, with balance regularization added to the objective,
the only difference as opposed to sparse autoencoder is the gradient
computation of unknown parameters $\theta_{e}$ and $\theta_{d}^{L},\theta_{d}^{R}$.
We clarify these differences in the following equations:

\begin{equation}
\begin{split}\nabla_{W_{e}^j}E= & \frac{\partial{J^{L}}}{\partial{W_{e}^j}}+\frac{\partial{J^{R}}}{\partial{W_{e}^j}}+\gamma(J^{L}-J^{R})(\frac{\partial{J^{L}}}{\partial{W_{e}^j}}-\frac{\partial{J^{R}}}{\partial{W_{e}^j}})\\
\nabla_{b_{e}^j}E= & \frac{\partial{J^{L}}}{\partial{b_{e}^j}}+\frac{\partial{J^{R}}}{\partial{b_{e}^j}}+\gamma(J^{L}-J^{R})(\frac{\partial{J^{L}}}{\partial{b_{e}^j}}-\frac{\partial{J^{R}}}{\partial{b_{e}^j}})
\end{split}
\end{equation}
and 
\begin{equation}
\begin{split} & \nabla_{W_{d}^{L}}E=\frac{\partial{J^{L}}}{\partial{W_{d}^{L}}}+\gamma(J^{L}-J^{R})\frac{\partial{J^{L}}}{\partial{W_{d}^{L}}}\\
 & \nabla_{b_{d}^{L}}E=\frac{\partial{J^{L}}}{\partial{b_{d}^{L}}}+\gamma(J^{L}-J^{R})\frac{\partial{J^{L}}}{\partial{b_{d}^{L}}}\\
 & \nabla_{W_{d}^{R}}E=\frac{\partial{J^{R}}}{\partial{W_{d}^{R}}}+\gamma(J^{L}-J^{R})(-\frac{\partial{J^{R}}}{\partial{W_{d}^{R}}})\\
 & \nabla_{b_{d}^{R}}E=\frac{\partial{J^{R}}}{\partial{b_{d}^{R}}}+\gamma(J^{L}-J^{R})(-\frac{\partial{J^{R}}}{\partial{b_{d}^{R}}})
\end{split}
\end{equation}

The exact form of gradients of $\theta_{e}$ and $\theta_{d}^{L},\theta_{d}^{R}$
varies according to different sparsity regularization $\Theta$ used
in the framework.

\section{Generating Synthetic Data}

Synthetic data are created to highlight the potential useful pattern
in real images.  In the proposed approach, the synthetic data are represented as a
parametric model of a set of control points and edges associated to
these points in the images. From the control points, the synthetic
images could be generated to simulate the real images in terms of
having the same structure or a similar appearance. Initially, the control
points are selected from a centralized prototype that generalize all images in
the same class. Then the locations of the control points are iteratively
optimized until convergence in order to minimize the distance between
synthetic images generated by control points and the real image. We
annotate the control points and edges associated to them as $\textbf{S}=\{\textbf{C},\textbf{E}\}$,
where $\textbf{C}=\{c_{i}\}_{i=1}^{n}$ is the set of the control points, and
$\textbf{E}=\{(c_{i},c_{j})\},1\leq i,j\leq n$ is the set of edges
connecting control points. A generalized algorithm of getting the
best matching synthetic image is provided in Algorithm \ref{Alg: MatchingSynthetic}.

\begin{algorithm}[htb]  
\small
 	\caption{ Get Matching Synthetic Image.}   
	\label{Alg: MatchingSynthetic}    	
\begin{algorithmic}[1]  	\REQUIRE ~~\\ 		
$\bullet$ A real image $U$. \\ 	
	$\bullet$ A set of control points 
$\textbf{S}=\{\textbf{C}, \textbf{E}\}$ 
with all control points $c_i\in \textbf{C}$ set to their initial positions.\\ 	
	$\bullet$ A prototype image $V$ generated using the initial $\textbf{S}$.  \\ 	
	\WHILE{\textbf{S} is not converged} 	
		\STATE \textbf{S} = OptimizeControlPoints($U, V, \textbf{S}$). 		
	\STATE Generate $V$ using $\textbf{S}$. 		
\ENDWHILE 
		\STATE Generate synthetic image $I$ using \textbf{S}. 	
	\RETURN $I$. 	
\end{algorithmic} 
 \end{algorithm}  

\subsection{Learning Synthetic Prototype from Data }

\begin{figure*}[ht]
\centering %
\begin{tabular}{cccccccccc}
\resizebox{0.07\textwidth}{!}{\rotatebox{0}{ \includegraphics{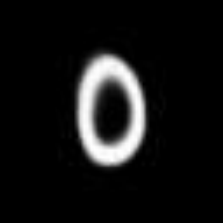}}}  
& 
\resizebox{0.07\textwidth}{!}{\rotatebox{0}{ \includegraphics{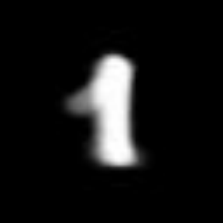}}}  
& 
\resizebox{0.07\textwidth}{!}{\rotatebox{0}{ \includegraphics{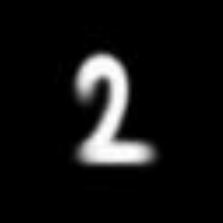}}}  
& 
\resizebox{0.07\textwidth}{!}{\rotatebox{0}{ \includegraphics{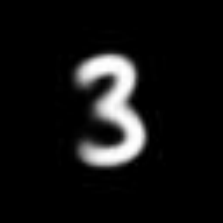}}}  
& 
\resizebox{0.07\textwidth}{!}{\rotatebox{0}{ \includegraphics{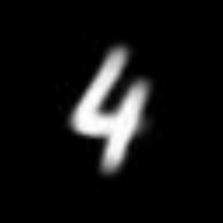}}}  
& 
\resizebox{0.07\textwidth}{!}{\rotatebox{0}{ \includegraphics{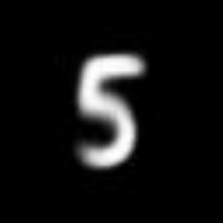}}} 
& 
\resizebox{0.07\textwidth}{!}{\rotatebox{0}{ \includegraphics{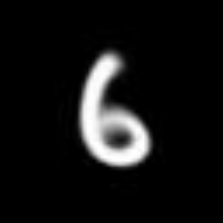}}} 
& 
\resizebox{0.07\textwidth}{!}{\rotatebox{0}{ \includegraphics{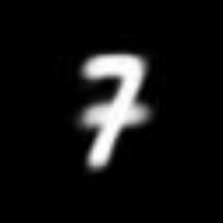}}} 
& 
\resizebox{0.07\textwidth}{!}{\rotatebox{0}{ \includegraphics{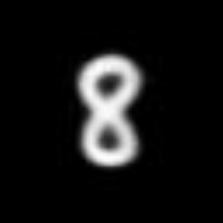}}} 
& 
\resizebox{0.07\textwidth}{!}{\rotatebox{0}{ \includegraphics{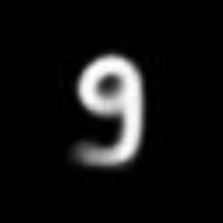}}} 
\\
\end{tabular} 
\caption{Illustration of average images of each digit after congealing.}
\label{fig: congealresults} 
\end{figure*}

%In some cases, the structure of data in an image is
%too complicated to manually design a uniform synthetic prototype. A typical
%example is given by the hand written digits recognition problem where a simple prototype is not going to generalize 
%various types of digits written by different people. 
%
%Therefore we learn the synthetic prototypes from the data and a digit prototype is generated for all images
In hand written digit dataset used in this work, we learn a centralized prototype from given data. 
A digit prototype is generated for all images with the same digit. Congealing
algorithm proposed in \cite{ME:00} is employed in this step to produce the synthetic prototypes for digits. In congealing,
the project transformations are applied to images to minimize a joint
entropy. Thus the prototype is considered to be an average image of all
images after congealing, shown in Fig. \ref{fig: congealresults}.

Then control points are evenly sampled from the boundary detected
from the prototype image. The control points needs to be mapped
to each digit image in order to generate a synthetic image. To find this mapping we implement an
approach that migrates the control points from the prototype images to destination image.

\begin{figure}[ht]
\centering %
\begin{tabular}{c}
\resizebox{0.3\textwidth}{!}{\rotatebox{0}{ \includegraphics{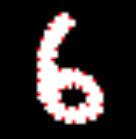}}}  
\\
\end{tabular} 
\caption{Illustration of control points on a digit image.}
\label{fig: controlpoints} 
\end{figure}

\begin{figure*}[ht]
\centering %
\begin{tabular}{ccccccc}
\resizebox{0.1\textwidth}{!}{\rotatebox{0}{ \includegraphics{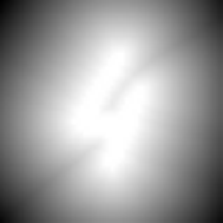}}}  
& 
\resizebox{0.1\textwidth}{!}{\rotatebox{0}{ \includegraphics{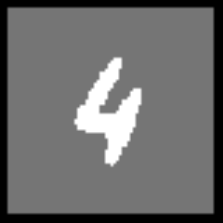}}}  
& 
\resizebox{0.1\textwidth}{!}{\rotatebox{0}{ \includegraphics{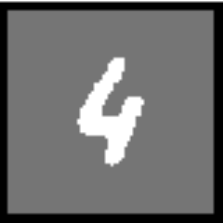}}}  
& 
\resizebox{0.1\textwidth}{!}{\rotatebox{0}{ \includegraphics{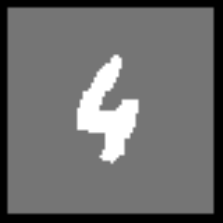}}}  
& 
\resizebox{0.1\textwidth}{!}{\rotatebox{0}{ \includegraphics{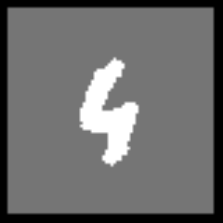}}}  
& 
\resizebox{0.1\textwidth}{!}{\rotatebox{0}{ \includegraphics{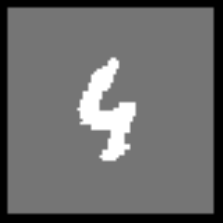}}} 
& 
\resizebox{0.1\textwidth}{!}{\rotatebox{0}{ \includegraphics{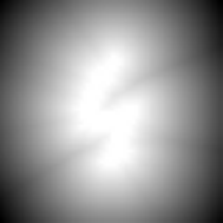}}} 
\\
\end{tabular} 
\caption{Illustrations of the migration of control points and intermediate
synthetic images generated using control points in each step. The
distance transform images of the synthetic prototype and real images
are shown as the left most and right most images respectively.}
\label{fig: migration1} 
\end{figure*}

\begin{figure}[tph]
\centering %
\begin{tabular}{c}
\resizebox{0.3\textwidth}{!}{\rotatebox{0}{ \includegraphics{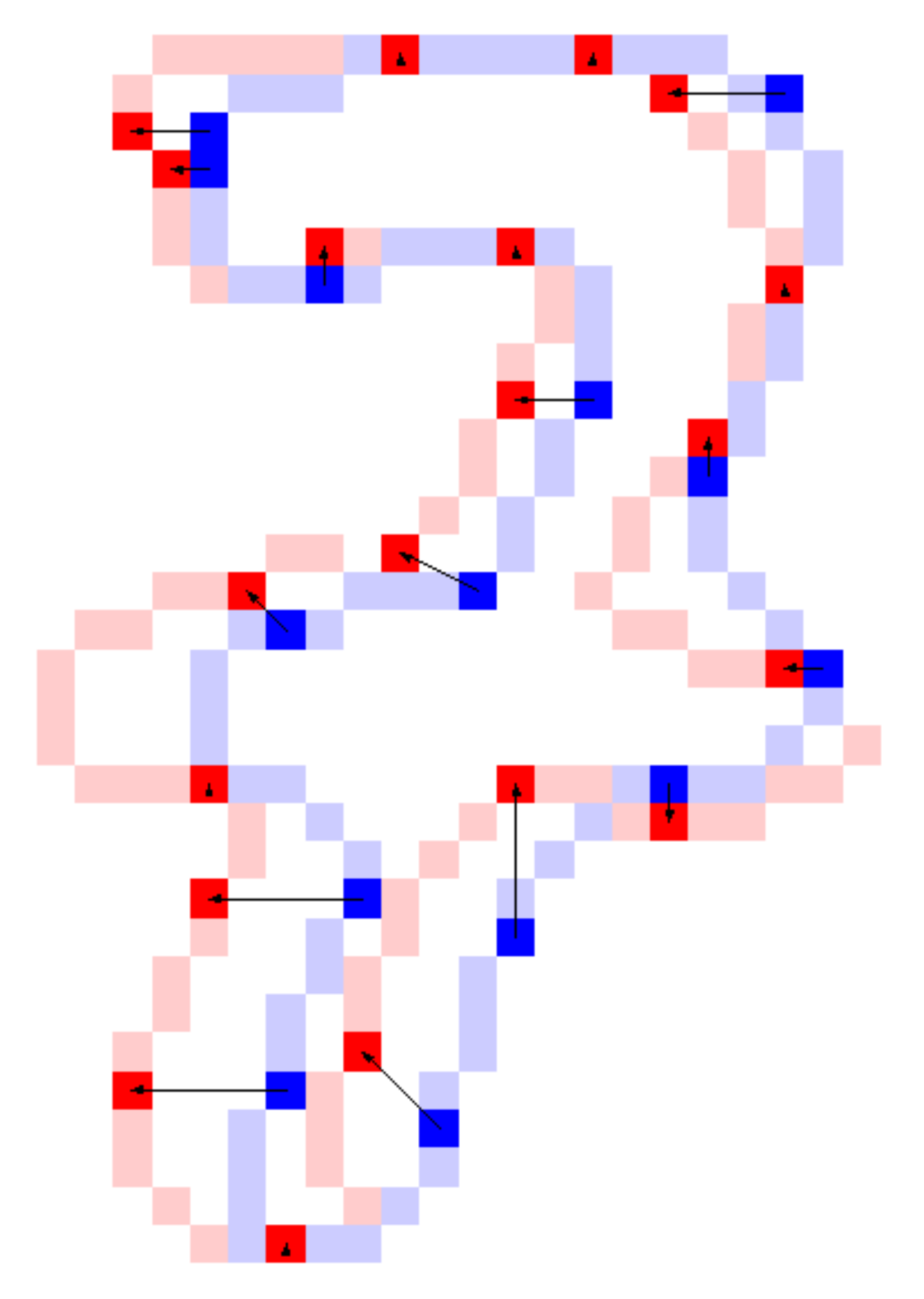}}} \tabularnewline
\end{tabular}\protect\protect\protect\caption{An example of migration of the control points from source image (blue)
to destination image (red).}

\label{fig: migration2} 
\end{figure}

This point migration algorithm is based on a series of intermediate
images generated in between synthetic prototype and destination image. To
generate the intermediate images, we binarize all the images and the
distance transformed images\cite{GB:86} of the synthetic prototype
and the real image are generated. Given the number of steps, an intermediate
image then is generated as a binarized image of linear interpolation
between two distance transformed images. In each
step, the control points are snapped to the closest boundary pixels
of the intermediate image. The algorithm of OptimizeControlPoints($U,V,\textbf{S}$)
in this situation is given in Algorithm \ref{Alg: OCP2}, we fix the
number of steps to $5$ in this algorithm. A step by step examples is given in Fig. \ref{fig: migration1}. A zoom in example showing how control points moved from one digit to another is shown in Fig. \ref{fig: migration2}.

\begin{algorithm}[htb]    
\small
	\caption{ OptimizeControlPoints($U, V, \textbf{S}$)}  
 	\label{Alg: OCP2}    
	\begin{algorithmic}[1]  		\REQUIRE ~~\\ 	
	$\bullet$ A real image $U$. \\ 	
	$\bullet$ A prototype of the synthetic image $\textbf{S}=\{\textbf{C}, \textbf{E}\}$.\\ 	
	$\bullet$ A synthetic image $V$. 
\\ 	
	\STATE $steps=10$. 	
	\STATE Compute distance transform image of $U, V$ as $U', V'$. 	
	\FOR{$i=1$ to $steps$} 			
\STATE $I=(1-\frac{i}{steps})U'+\frac{i}{steps}V'$.\\
			\STATE $I$=Binarize($I$).\\ 	
		\STATE Update $\textbf{S}$ by snapping to the closest boundary pixel on $I$.\\ 	
	\ENDFOR 		\STATE Set the status of $\textbf{S}$ to be converged. 	
	\RETURN \textbf{S}. 	
\end{algorithmic} 
 \end{algorithm} 

\begin{figure*}[ht]
\centering %
\begin{tabular}{ccccc}
\resizebox{0.15\textwidth}{!}{\rotatebox{0}{ \includegraphics{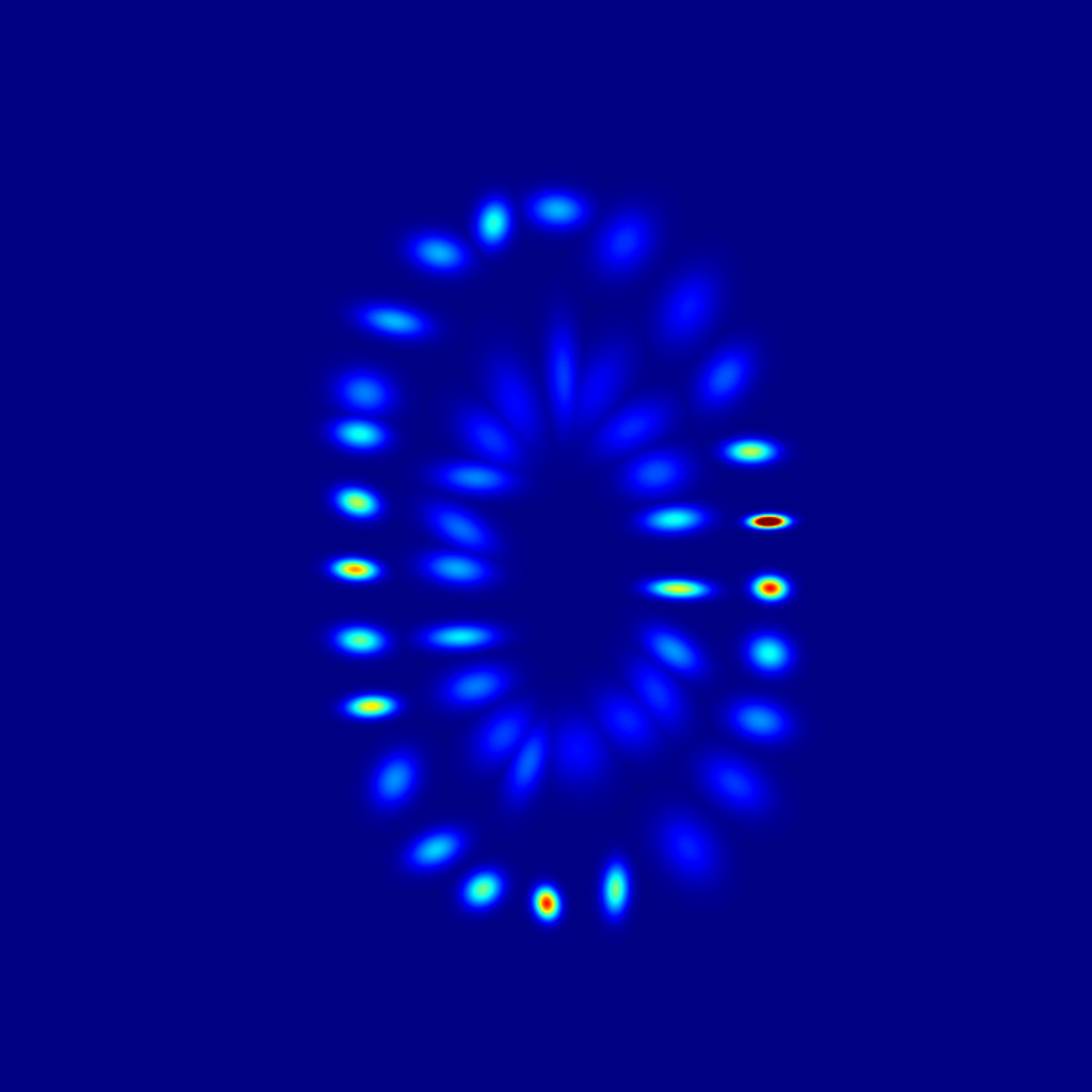}}}  
& 
\resizebox{0.15\textwidth}{!}{\rotatebox{0}{ \includegraphics{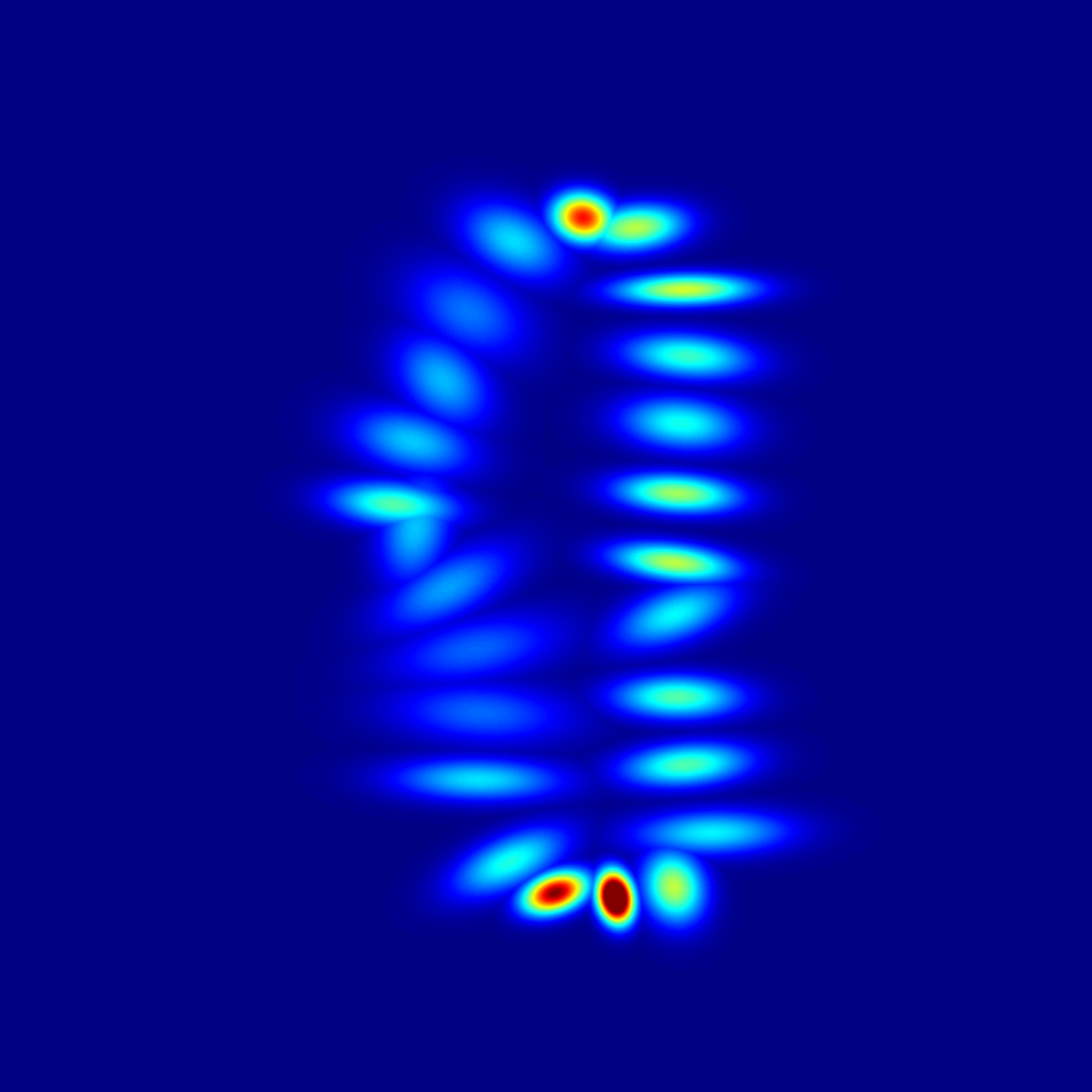}}}  
& 
\resizebox{0.15\textwidth}{!}{\rotatebox{0}{ \includegraphics{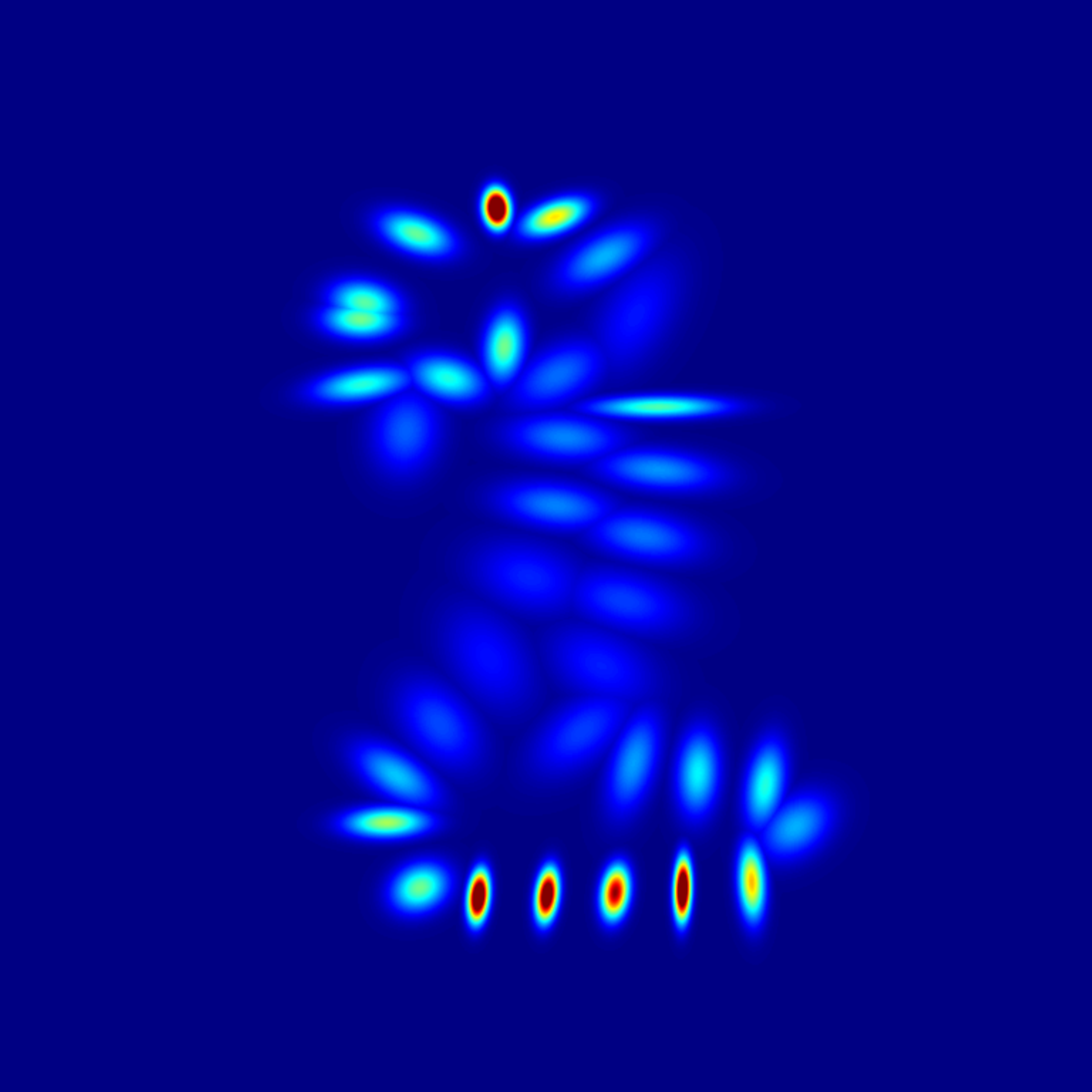}}}  
& 
\resizebox{0.15\textwidth}{!}{\rotatebox{0}{ \includegraphics{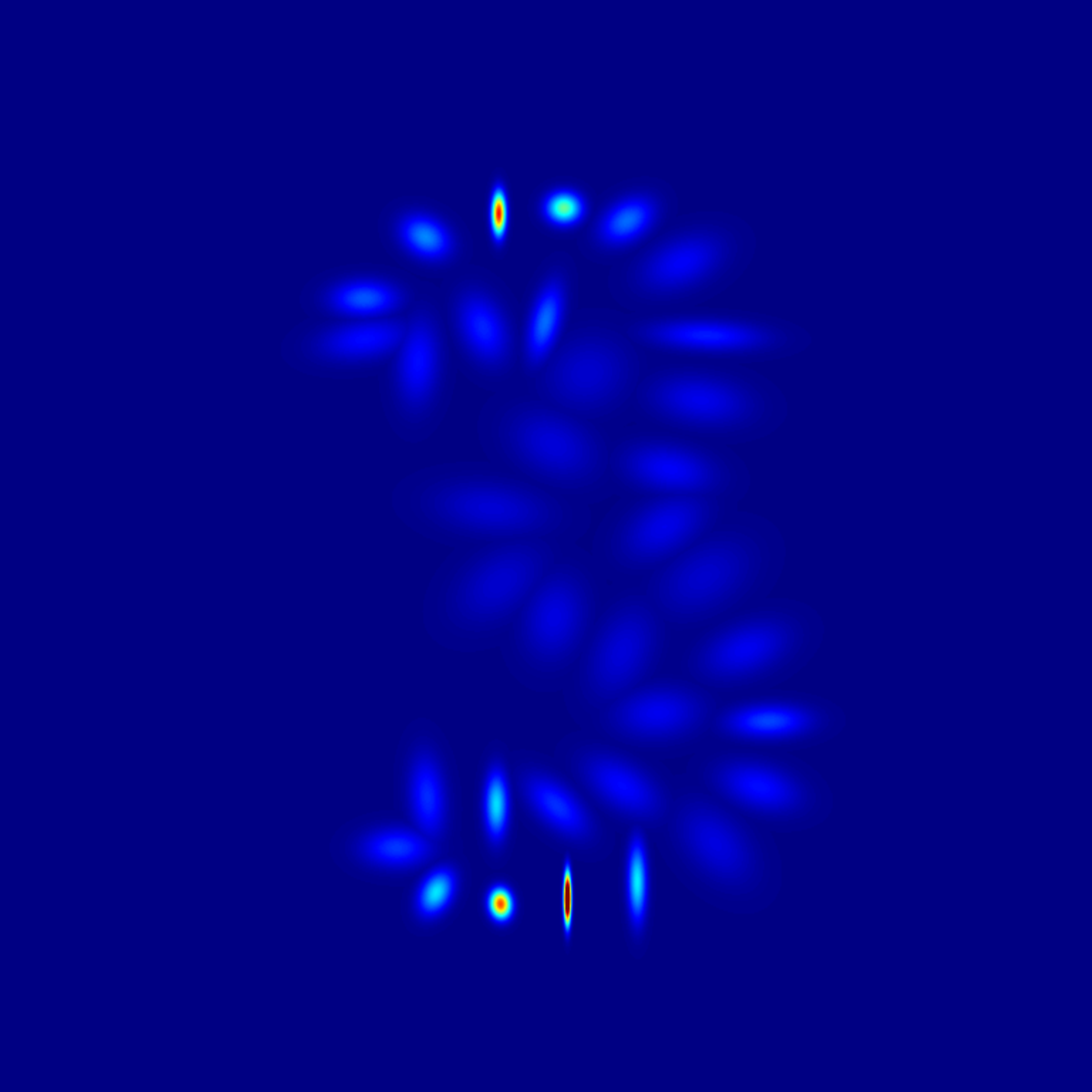}}}  
& 
\resizebox{0.15\textwidth}{!}{\rotatebox{0}{ \includegraphics{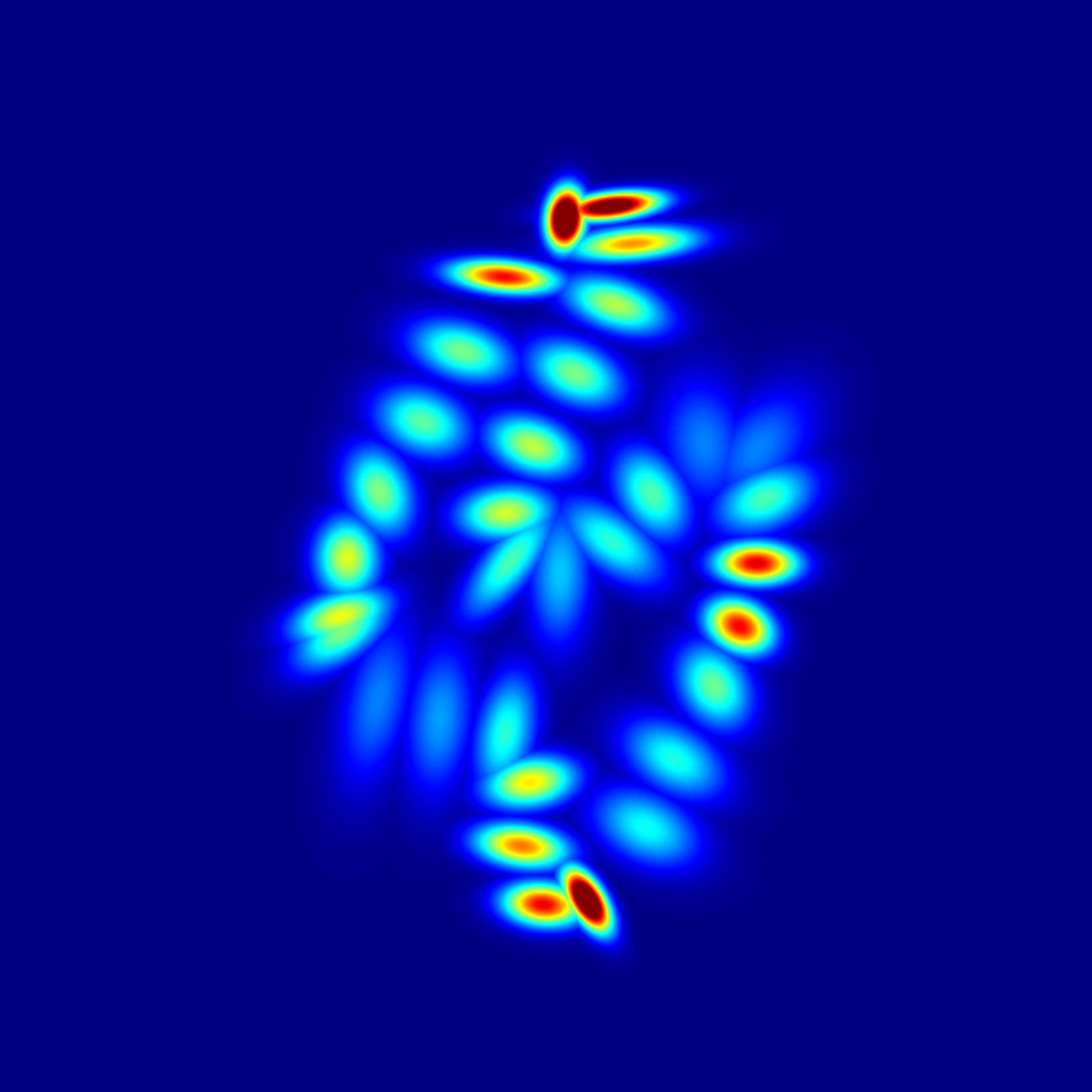}}}  
\\
\resizebox{0.15\textwidth}{!}{\rotatebox{0}{ \includegraphics{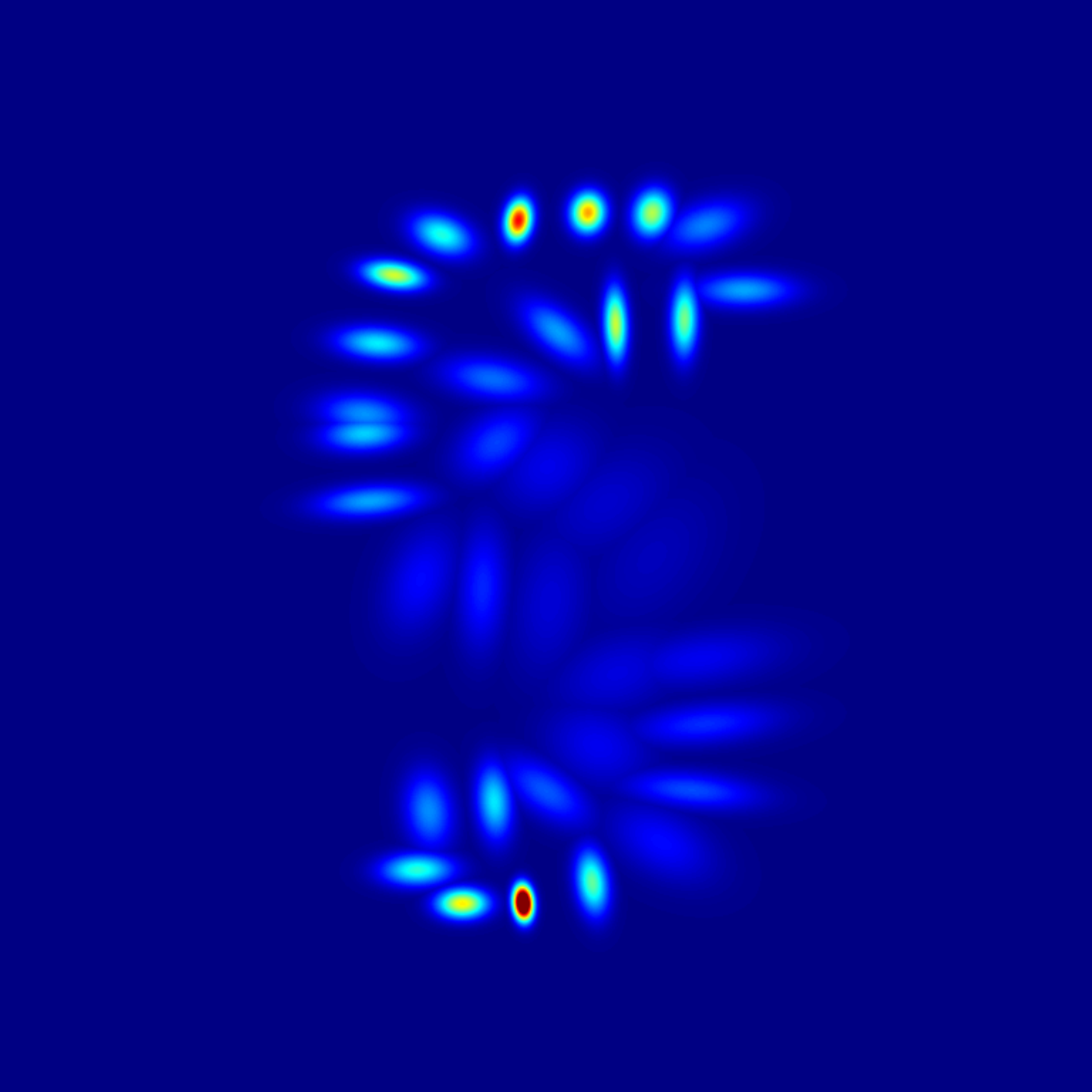}}} 
& 
\resizebox{0.15\textwidth}{!}{\rotatebox{0}{ \includegraphics{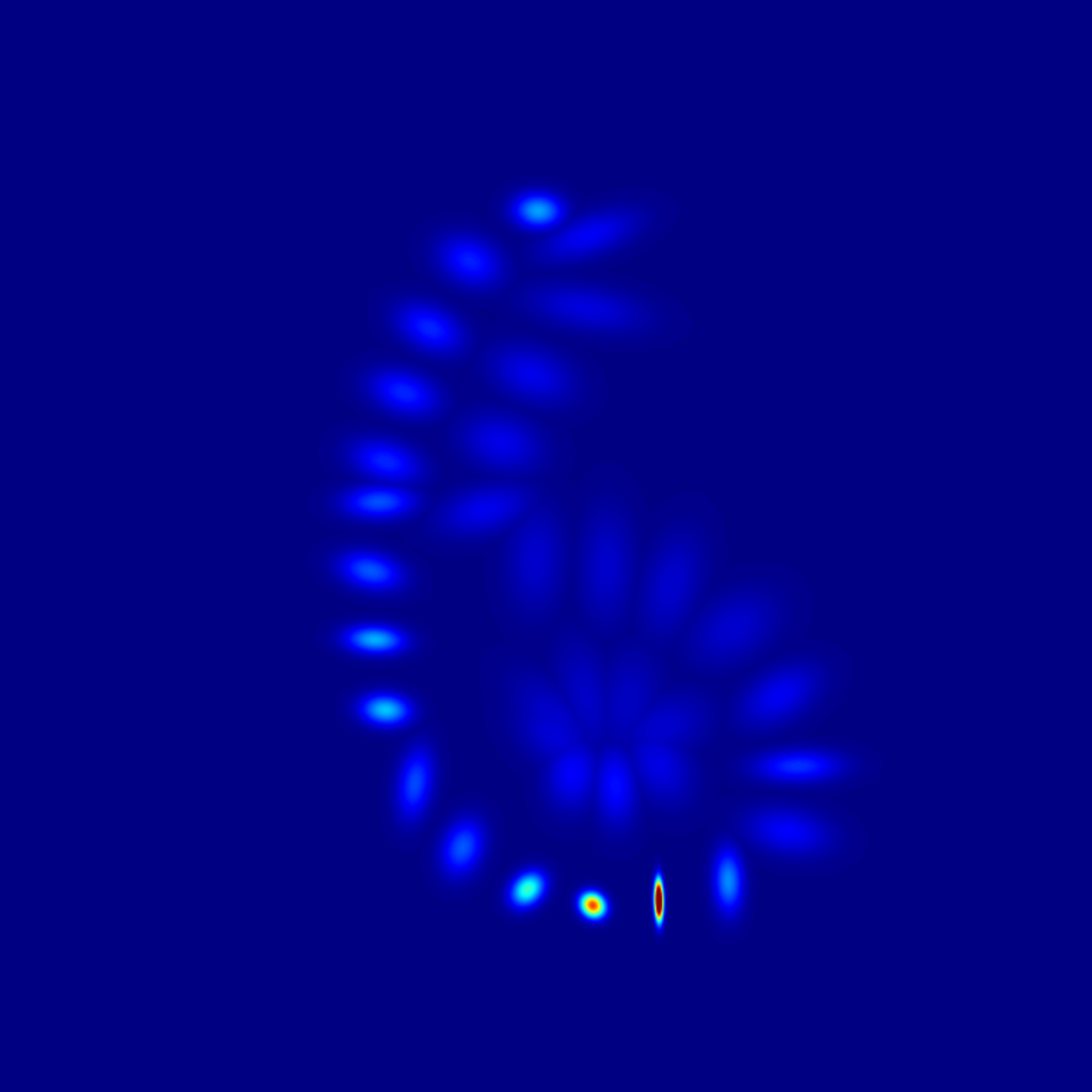}}} 
& 
\resizebox{0.15\textwidth}{!}{\rotatebox{0}{ \includegraphics{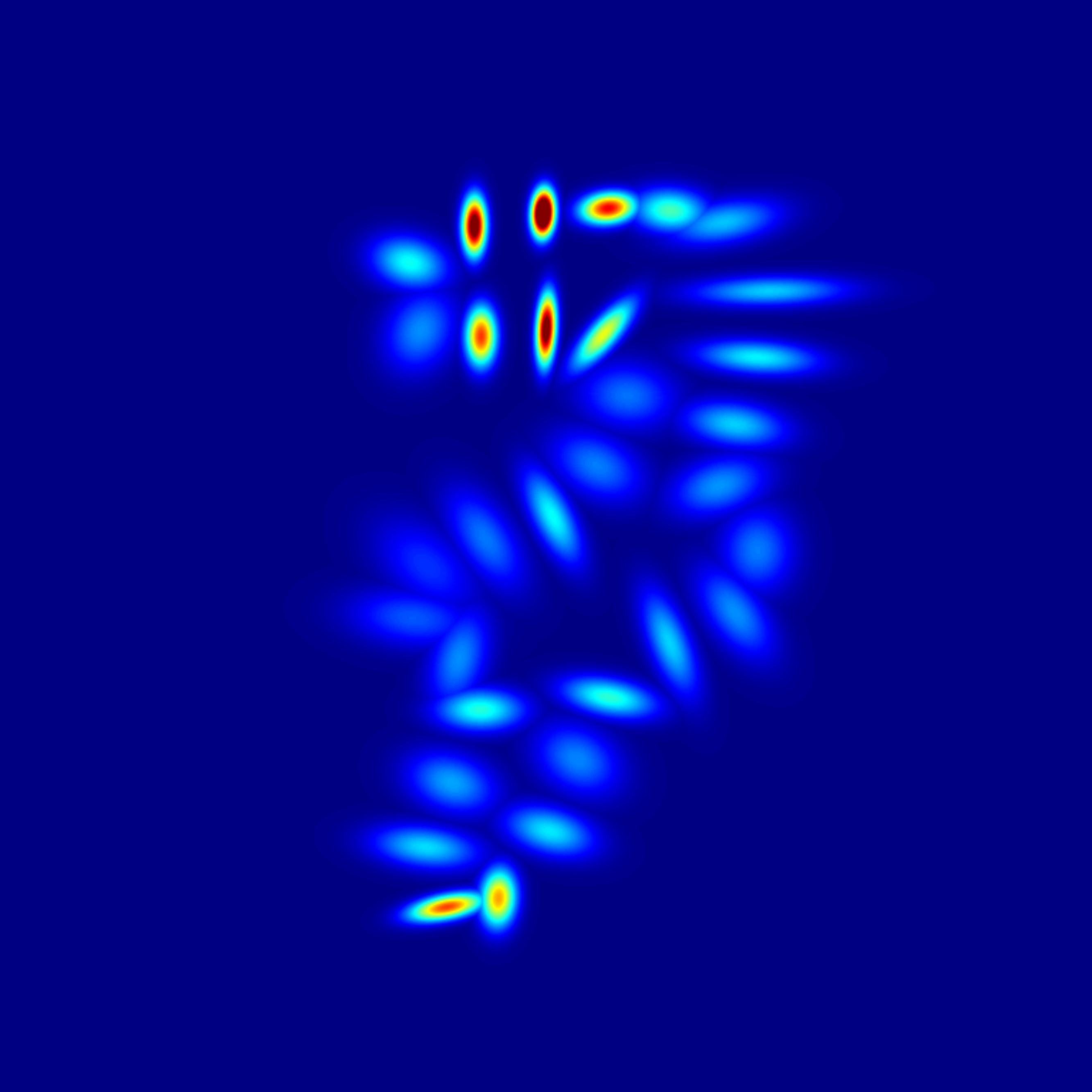}}} 
& 
\resizebox{0.15\textwidth}{!}{\rotatebox{0}{ \includegraphics{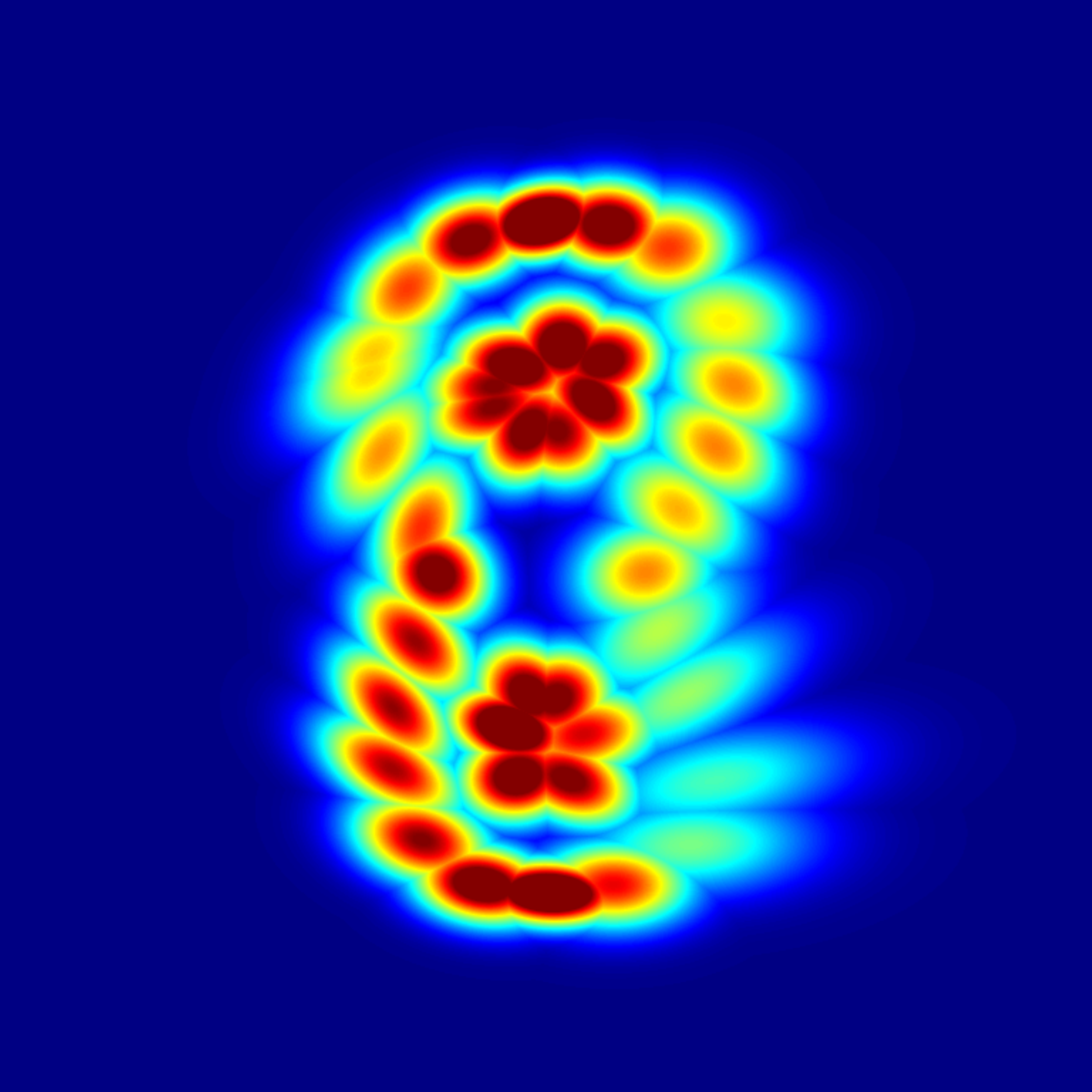}}} 
& 
\resizebox{0.15\textwidth}{!}{\rotatebox{0}{ \includegraphics{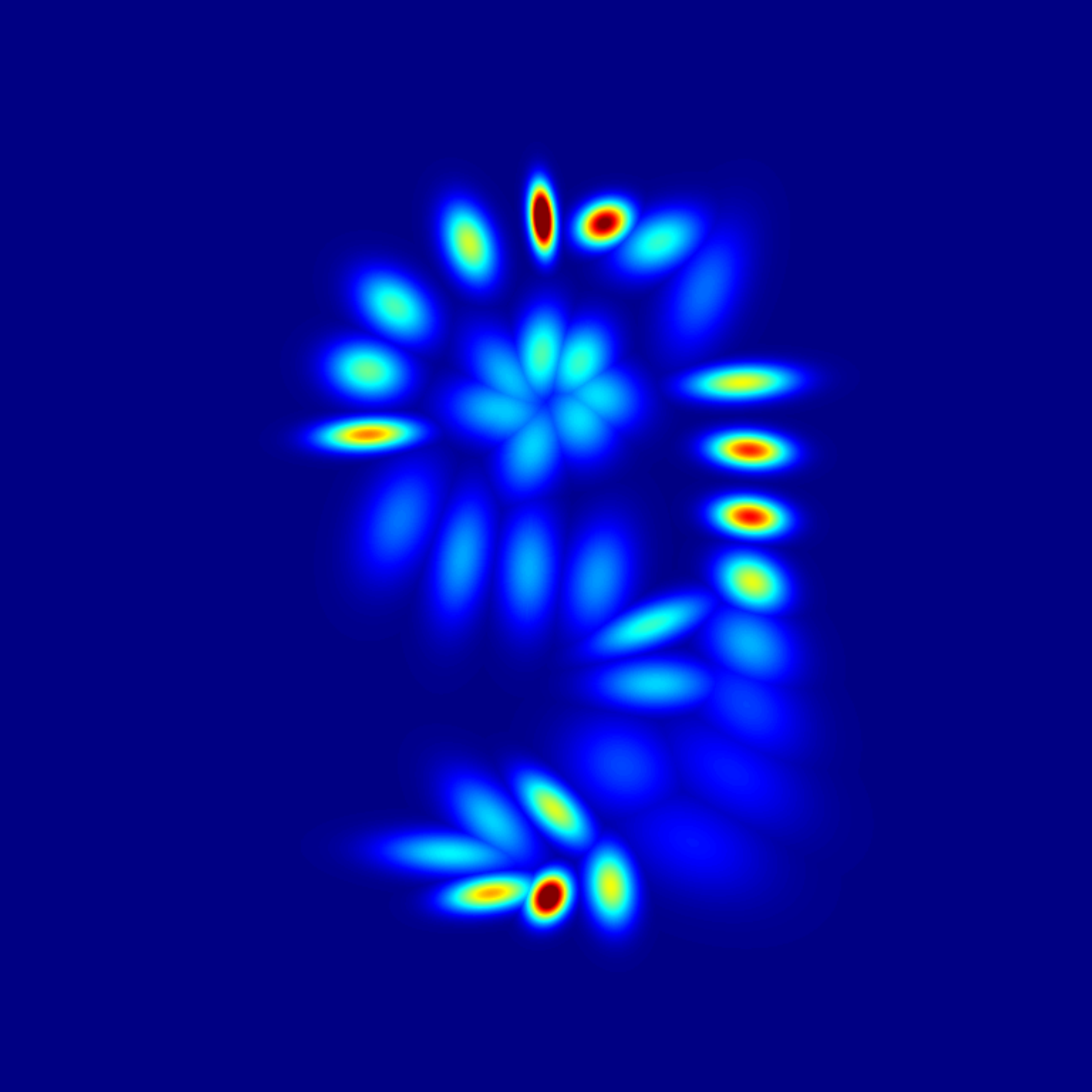}}} 
\\
\end{tabular} 
\caption{Illustration of distributions of control points on each digit image, where colors from blue to red are used to represent the probability density from low to high.}
\label{fig: controlpointsdistribution} 
\end{figure*}

To generate more synthetic digit images, We assume the distribution of control points on each digit image follows a multivariant normal distribution that $C\sim N(\mu, \Sigma)$ where $\mu$ and $\Sigma$ are computed using existing control points. The visualization of the distribution of control points of each digit is then shown in Fig. \ref{fig: controlpointsdistribution}.

\end{document}